\documentclass[twoside,11pt]{article}

\usepackage{jmlr2e}
\usepackage{times}
\usepackage{graphicx} 
\usepackage{subfigure}

\usepackage{natbib}

\usepackage{algorithm}
\usepackage{algorithmic}
\usepackage{amsmath,bm}
\usepackage{amssymb, multirow, paralist}
\usepackage{graphicx,color}

\def \R {\mathbb{R}}
\def \e {\epsilon}

\def \x {\mathbf{x}}
\def \O {\mathcal{O}}
\def \X {\mathbf{X}}
\def \M {\widehat{M}_*}

\def \Gh {\widehat{G}}
\def \u {\mathbf{u}}
\def \v {\mathbf{v}}
\def \z {\mathbf{z}}
\def \xt {\widetilde{\x}}
\def \Zt {\widetilde{Z}}
\def \zt {\widetilde{\z}}
\def \ah {\widehat{\bm \alpha}}
\def \p {\mathbf{p}}
\def \N {\mathcal{N}}

\def \as {{\bm \alpha}_*}
\def \D {\mathcal{D}}
\def \xh {\widehat{\x}}
\def \Ah {\widehat{A}}

\def \q {\mathbf{q}}

\newtheorem{thm}{Theorem}

\newtheorem{cor}[thm]{Corollary}

\jmlrheading{}{}{}{}{}{Qi Qian, Rong Jin, Lijun Zhang and Shenghuo Zhu}

\ShortHeadings{Towards Making High Dimensional Distance Metric Learning Practical}{Qi Qian, Rong Jin, Lijun Zhang and Shenghuo Zhu}
\firstpageno{1}

\begin{document}

\title{Towards Making High Dimensional Distance Metric Learning Practical}

\author{\name Qi Qian \email qianqi@cse.msu.edu \\
        \name Rong Jin \email rongjin@cse.msu.edu \\
        \name Lijun Zhang \email zlj@cse.msu.edu \\
        \addr Department of Computer Science and Engineering\\
        Michigan State University\\
        East Lansing, MI, 48824, USA
       \AND
       \name Shenghuo Zhu \email zsh@nec-labs.com \\
       \addr NEC Laboratories America\\
        Cupertino, CA, 95014, USA}

\editor{}

\maketitle

\begin{abstract}

In this work, we study distance metric learning (DML) for high dimensional data. A typical approach for DML with high dimensional data is to perform the dimensionality reduction first before learning the distance metric. The main shortcoming of this approach is that it may result in a suboptimal solution due to the subspace removed by the dimensionality reduction method. In this work, we present a dual random projection frame for DML with high dimensional data that explicitly addresses the limitation of dimensionality reduction for DML. The key idea is to first project all the data points into a low dimensional space by random projection, and compute the dual variables using the projected vectors. It then reconstructs the distance metric in the original space using the estimated dual variables. The proposed method, on one hand, enjoys the light computation of random projection, and on the other hand, alleviates the limitation of most dimensionality reduction methods. We verify both empirically and theoretically the effectiveness of the proposed algorithm for high dimensional DML.

\end{abstract}


\section{Introduction}


Distance metric learning (DML) is essential to many machine learning tasks, including ranking~\citep{ChechikSSB10,Lim13}, $k$-nearest neighbor ($k$-NN) classification~\citep{weinberger2009} and $k$-means clustering~\citep{XingNJR02}. It finds a good metric by minimizing the distance between data pairs in the same classes and maximizing the distance between data pairs from different classes~\citep{XingNJR02,GlobersonR05,liu2006,DavisKJSD07, weinberger2009, shaw2011}. The main computational challenge of DML arises from the constraint that the learned matrix has to be positive semi-definite (PSD). It is computationally demanding even with a stochastic gradient descent (SGD) because it has to project the intermediate solutions onto the PSD cone at every iteration. In a recent study~\citep{ChechikSSB10}, the authors show empirically that it is possible to learn a good distance metric using online learning without having to perform the projection at each iteration. In fact, only one projection into the PSD cone is performed at the end of online learning to ensure that the resulting matrix is PSD~\footnote{We note that this is different from the algorithms presented in~\citep{hazan2012,mahdavi2012}. Although these two algorithms only need either one or no projection step, they introduce additional mechanisms to prevent the intermediate solutions from being too far away from the PSD cone, which could result in a significant overhead per iteration.}. Our study of DML follows the same paradigm, to which we refer as {\it one-projection paradigm}.

Although the one-projection paradigm resolves the computational challenge from projection onto the PSD cone, it still suffers from a high computational cost when each data point is described by a large number of features. This is because, for $d$ dimensional data points, the size of learned matrix will be $\O(d^2)$, and as a result, the cost of computing the gradient, the fundamental operation for any first order optimization method, will also be $\O(d^2)$. The focus of this work is to develop an efficient first order optimization method for high dimensional DML that avoids $\O(d^2)$ computational cost per iteration.


Several approaches have been proposed to reduce the computation cost for high dimensional DML. In~\citep{DavisD08}, the authors assume that the learned metric $M$ is of low rank, and write it as $M = LL^{\top}$, where $L\in \R^{d\times r}$ with $r \ll d$. Instead of learning $M$, the authors proposed to learn $L$ directly, which reduces the cost of computing the gradient from $\O(d^2)$ to $\O(dr)$. A similar idea was studied in~\citep{weinberger2009}. The main problem with this approach is that it will result in non-convex optimization. An alternative approach is to reduce the dimensionality of data using dimensionality reduction methods such as principal component analysis (PCA)~\citep{weinberger2009} or random projection (RP)~\citep{TsagkatakisS10}. Although RP is computationally more efficient than PCA, it often yields significantly worse performance than PCA unless the number of random projections is sufficiently large~\citep{FradkinM03}. We note that although RP has been successfully applied to many machine learning tasks, e.g., classification~\citep{RahimiR07}, clustering~\citep{BoutsidisZD10} and regression~\citep{Maillard12}, only a few studies examined the application of RP to DML, and most of them with limited success.


In this paper, we propose a {\it dual random projection} approach for high dimensional DML. Our approach, on one hand, enjoys the light computation of random projection, and on the other hand, significantly improves the effectiveness of random projection. The main limitation of using random projection for DML is that all the columns/rows of the learned metric will lie in the subspace spanned by the random vectors. We address this limitation of random projection by
\begin{compactitem}
\item first estimating the {\it dual} variables based on the random projected vectors and,
\item then reconstructing the distance metric using the estimated dual variables and data vectors in the original space.
\end{compactitem}
Since the final distance metric is computed using the original vectors, not the randomly projected vectors, the column/row space of the learned metric will NOT be restricted to the subspace spanned by the random projection, thus alleviating the limitation of random projection. We verify the effectiveness of the proposed algorithms both empirically and theoretically.

We finally note that our work is built upon the recent work~\citep{Zhang13} on random projection where a dual random projection algorithm is developed for linear classification. Our work differs from~\citep{Zhang13} in that we apply the theory of dual random projection to DML. More importantly, we have made an important progress in advancing the theory of dual random projection. Unlike the theory in~\citep{Zhang13} where the data matrix is assumed to be low rank or approximately low rank, our new theory of dual random projection is applicable to any data matrix even when it is {\it NOT} approximately low rank. This new analysis significantly broadens the application domains where dual random projection is applicable, which is further verified by our empirical study.


The rest of the paper is organized as follows: Section~\ref{sec:related} introduces the methods that are related to the proposed method. Section~\ref{sec:method} describes the proposed dual random projection approach for DML and the detailed algorithm for solving the dual problem in the subspace spanned by random projection. Section~\ref{sec:exp} summarizes the results of the empirical study, and Section~\ref{sec:conclusion} concludes this work with future directions.

\section{Related Work}
\label{sec:related}


Many algorithms have been developed for DML~\citep{XingNJR02,GlobersonR05,DavisKJSD07, weinberger2009}. Exemplar DML algorithms are MCML~\citep{GlobersonR05}, ITML~\citep{DavisKJSD07}, LMNN~\citep{weinberger2009} and OASIS~\citep{ChechikSSB10}. Besides algorithms, several studies were devoted to analyzing the generalization performance of DML~\citep{JinWZ09, Bellet13}. Survey papers~\citep{liu2006,Kulis13} provide detailed investigation about the topic. Although numerous studies were devoted to DML, only a limited progress is made to address the high dimensional challenge in DML~\citep{DavisD08,weinberger2009,QiTZCZ09,Lim13}. In~\citep{DavisD08,weinberger2009}, the authors address the challenge of high dimensionality by enforcing the distance metric to be a low rank matrix. \citep{QiTZCZ09,Lim13} alleviate the challenge of learning a distance metric $M$ from high dimensional data by assuming $M$ to be a sparse matrix. The main shortcoming of these approaches is that they have to place strong assumption on the learned metric, significantly limiting their application. In addition, these approaches will result in non-convex optimization problems that are usually difficult to solve. In contrast, the proposed DML algorithm does not have to make strong assumption regarding the learned metric.



Random projection is widely used for dimension reduction in various learning tasks~\citep{RahimiR07, BoutsidisZD10,Maillard12}. Unfortunately, it requires a large amount of random projections for the desired result~\citep{FradkinM03}, and this limits its application in DML, where the computational cost is proportion to the square of dimensions. Dual random projection is first introduced for linear classification task~\citep{Zhang13} and following aspects make our work significantly different from the initial study~\citep{Zhang13}: First, we apply dual random projection for DML, where the number of variables is quadratic to the dimension and the dimension crisis is more serious than linear classifier. Second, we optimize the dual problem directly rather than the primal problem in the subspace as the previous work. Consequently, non-smoothed loss (e.g., hinge loss) could be used for the proposed method. Last, we give the theoretical guarantee when the dataset is not low rank, which is an important assumption for the study~\citep{Zhang13}. All of these efforts try to efficiently learn a distance metric for high dimensional datasets and sufficient empirical study verifies the success of our method.

\section{Dual Random Projection for Distance Metric Learning}
\label{sec:method}

Let $\X = (\x_1,\cdots,\x_n) \in \R^{d\times n}$ denote the collection of training examples. Given a PSD matrix $M$, the distance between two examples $\x_i$ and $\x_j$ is given as
\begin{eqnarray*}
 d_M(\x_i, \x_j) = (\x_i-\x_j)^\top M(\x_i-\x_j).
\end{eqnarray*}
The proposed framework for DML will be based on triplet constraints, not pairwise constraints. This is because several previous studies have suggested that triplet constraints are more effective than pairwise constraints~\citep{weinberger2009,ChechikSSB10,shaw2011}. Let $\mathcal{D} = \{(\x_i^1, \x_j^1, \x_k^1), \ldots, (\x_i^N, \x_j^N, \x_k^N)\}$ be the set of triplet constraints used for training, where $\x^t_i$ is expected to be more similar to $\x^t_j$ than to $\x^t_k$. Our goal is to learn a metric function $M$ that is consistent with most of the triplet constraints in $\mathcal{D}$, i.e.
\begin{eqnarray*}
\forall (\x_i^t, \x_j^t, \x_k^t) \in \D ,\ \ (\x^t_i\!-\!\x^t_j)^\top M(\x^t_i\!-\!\x^t_j)\! +\! 1 \leq (\x^t_i\!-\!\x^t_k)^\top M(\x^t_i\!-\!\x^t_k)
\end{eqnarray*}
Following the empirical risk minimization framework, we cast the triplet constraints based DML into the following optimization problem:
\begin{eqnarray}
\min\limits_{M \in S_{d}} \frac{\lambda}{2}\|M\|_F^2+\frac{1}{N}\sum_{t=1}^N\ell(\langle M,A_t\rangle) \label{eqn:primal}
\end{eqnarray}
where $S_d$ stands for the symmetric matrix of size $d\times d$, $\lambda > 0$ is the regularization parameter, $\ell(\cdot)$ is a convex loss function, $A_t = (\x_i^t-\x_k^t)(\x_i^t-\x_k^t)^\top-(\x_i^t-\x_j^t)(\x_i^t-\x_j^t)^\top$, and $\langle \cdot, \cdot \rangle$ stands for the dot product between two matrices. We note that we did not enforce $M$ in (\ref{eqn:primal}) to be PSD because we follow the one-projection paradigm proposed in~\citep{ChechikSSB10} that first learns a symmetric matrix $M$ by solving the optimization problem in (\ref{eqn:primal}) and then projects the learned matrix $M$ onto the PSD cone. We emphasize that unlike~\citep{Zhang13}, we did not assume $\ell(\cdot)$ to be smooth, making it possible to apply the proposed approach to the hinge loss.

Let $\ell_*(\cdot)$ be the convex conjugate of $\ell(\cdot)$. The dual problem of (\ref{eqn:primal}) is given by
\begin{eqnarray*}
\max\limits_{\alpha_1, \ldots, \alpha_N} -\frac{1}{N}\sum_{t=1}^N\ell_*(\alpha_t) - \frac{1}{2\lambda N^2}\left\|\sum_{t=1}^N \alpha_t A_t \right\|_F^2
\end{eqnarray*}
which is equivalent to
\begin{eqnarray}
\max\limits_{{\bm \alpha}\in [-1,0]^N} -\sum_{t=1}^N \ell_*(\alpha_t) - \frac{1}{2\lambda N}{\bm \alpha}^\top G {\bm \alpha} \label{eqn:dual}
\end{eqnarray}
where ${\bm \alpha} = (\alpha_1,\cdots,\alpha_N)^\top$ and $G = [G_{a,b}]_{N\times N}$ is a matrix of $N\times N$ with $G_{a,b} = \langle A_a, A_b \rangle$. We denote by $M_*\in \R^{d\times d}$ the optimal primal solution to (\ref{eqn:primal}), and by $\as\in\R^{N}$ the optimal dual solution to (\ref{eqn:dual}). Using the first order condition for optimality, we have
\begin{eqnarray}
M_* = - \frac{1}{\lambda N}\sum_{t=1}^N \alpha_*^tA_t \label{eqn:primal-dual}
\end{eqnarray}

\subsection{Dual Random Projection for Distance Metric Learning}

Directly solving the primal problem in (\ref{eqn:primal}) or the dual problem in (\ref{eqn:dual}) could be computational expensive when the data is of high dimension and the number of training triplets is very large. We address this challenge by inducing a random matrix $R\in \R^{d\times m}$, where $m \ll d$ and $R_{i,j}\sim \N(0, 1/m)$, and projecting all the data points into the low dimensional space using the random matrix, i.e., $\xh_i = R^\top \x_i$. As a result, $A_t$, after random projection, becomes $\Ah_t = R^{\top}A_tR$.

A typical approach of using random projection for DML is to obtain a matrix $M_s$ of size $m\times m$ by solving the primal problem with the randomly projected vectors $\{\xh_i\}_{i=1}^n$, i.e.
\begin{eqnarray}
\min\limits_{M \in S_{m}} \frac{\lambda}{2}\|M\|_F^2 + \frac{1}{N}\sum_{t=1}^N \ell(\langle M, \Ah_t\rangle ) \label{eqn:primal-1}
\end{eqnarray}
Given the learned metric $M_s$, for any two data points $\x$ and $\x'$, their distance is measured by $(\x - \x')^\top R M_s R^{\top}(\x - \x') = (\x- \x')^{\top}M'(\x - \x')$, where $M' = RM_s R^{\top} \in \R^{d\times d}$ is the effective metric in the original space $\R^d$. The key limitation of this random projection approach is that both the column and row space of $M'$ are restricted to the subspace spanned by vectors in random matrix $R$.

Instead of solving the primal problem, we proposed to solve the dual problem using the randomly projected data points $\{\xh_i\}_{i=1}^n$, i.e.
\begin{eqnarray}
 \max\limits_{{\bm \alpha} \in \R^N} -\sum_{t=1}^N \ell_*(\alpha_t) - \frac{1}{2\lambda N}{\bm \alpha}^\top \Gh {\bm \alpha} \label{eqn:dual2}
\end{eqnarray}
where $\Gh_{a,b} = \langle R^\top A_a R, R^\top A_b R\rangle$. After obtaining the optimal solution $\ah_*$ for (\ref{eqn:dual2}), we reconstruct the metric by using the dual variables $\ah_*$ and data matrix $X$ in the original space, i.e.
\vspace{-3mm}
\begin{eqnarray}
\M =  - \frac{1}{\lambda N}\sum_{t=1}^N \widehat{\alpha}^t_*A_t \label{eqn:recover}
\end{eqnarray}
It is important to note that unlike the random projection approach, the recovered metric $\M$ in (\ref{eqn:recover}) is not restricted by the subspace spanned by the random vectors, a key to the success of the proposed algorithm.

Alg.~\ref{alg:1} summarizes the key steps for the proposed dual random projection method for DML. Following one-projection paradigm~\citep{ChechikSSB10}, we project the learned symmetric matrix $M$ onto the PSD cone at the end of the algorithm. The key component of Alg.~\ref{alg:1} is to solve the optimization problem in (\ref{eqn:dual}) at Step 4 accurately. We choose stochastic dual coordinate ascent (SDCA) method for solving the dual problem (\ref{eqn:dual2}) because it enjoys a linear convergence when the loss function is smooth, and is shown empirically to be significantly faster than the other stochastic optimization methods~\citep{tzhang2012}. We use the combination strategy recommended in~\citep{tzhang2012}, denoted by {\bf CSDCA}, which uses SGD for the first epoch and then applies SDCA for the rest epochs.

\begin{algorithm}[t]
\caption{Dual Random Projection Method (DuRP) for DML}
\begin{algorithmic}[1]
\STATE {\bf Input}: the triplet constraints $\mathcal{D}$ and the number of random projections $m$.
\STATE Generate a random matrix $R\in \R^{d\times m}$ and $R_{i,j}\sim \N(0, 1/m)$.
\STATE Project each example as $\xh = R^{\top}\x$.
\STATE Solve the optimization problem (\ref{eqn:dual2}) and obtain the optimal solution $\ah_*$ \label{eqn:sub}
\STATE Recover the solution in the original space by $\M =  - \frac{1}{\lambda N}\sum_t \widehat{\alpha}^t_*A_t$
\STATE Output: $\Pi_{PSD}(\M)$
\end{algorithmic}\label{alg:1}
\end{algorithm}

\subsection{Main Theoretical Results}

First, similar to~\citep{Zhang13}, we consider the case when the data matrix $X$ is of low rank. The theorem below shows that under the low rank assumption, with a high probability, the distance metric recovered by Algorithm~\ref{alg:1} is nearly optimal.
\begin{thm} \label{thm:low-rank}
Let $M_*$ be the optimal solution to (\ref{eqn:primal}). Let $\ah_*$ be the optimal solution for (\ref{eqn:dual2}), and let $\M$ be the solution recovered from $\ah_*$ using (\ref{eqn:recover}). Under the assumption that all the data points lie in the subspace of $r$-dimension, for any $0<\varepsilon \leq 1/6$, with a probability at least $1 - \delta$, we have
\begin{eqnarray*}
   \|\Pi_{PSD}(M_*)-\Pi_{PSD}(\M)\|_F \leq \frac{3\varepsilon}{1 - 3\varepsilon} \|M_*\|_F
\end{eqnarray*}
provided $m \geq \frac{(r+1)\log(2r/\delta)}{c\varepsilon^2}$ and constant $c$ is at least $1/3$.
\end{thm}
The proof of Theorem~\ref{thm:low-rank} can be found in appendix. Theorem~\ref{thm:low-rank} indicates that if the number of random projections is sufficiently large (i.e. $m = \Omega(r\log r)$), we can recover the optimal solution in the original space with a small error. It is important to note that our analysis, unlike~\citep{Zhang13}, can be applied to non-smooth loss such as the hinge loss.

In the second case, we assume the loss function $\ell(\cdot)$ is $\gamma$-smooth (i.e., $|\ell'(z) - \ell'(z')| \leq \gamma |z - z'|$). The theorem below shows that the dual variables obtained by solving the optimization problem in (\ref{eqn:dual2}) can be close to the optimal dual variables, even when the data matrix $X$ is {\it NOT} low rank or approximately low rank. For the presentation of theorem, we first define a few important quantities. Define matrices $M^1 \in \R^{N\times N}$, $M^2 \in \R^{N\times N}$, $M^3 \in \R^{N\times N}$, and $M^4 \in \R^{N\times N}$ as
\begin{eqnarray*}
M^1_{a,b} & = & \|(\x_i^a - \x_k^a)\|_2^2\times\|(\x_i^b - \x_k^b)\|_2^2 \\
M^2_{a,b} & = & \|(\x_i^a - \x_j^a)\|_2^2\times\|(\x_i^b - \x_j^b)\|_2^2 \\
M^3_{a,b} & = & \|(\x_i^a - \x_k^a)\|_2^2\times\|(\x_i^b - \x_j^b)\|_2^2\\
M^4_{a,b} & = & \|(\x_i^a - \x_j^a)\|_2^2\times\|(\x_i^b - \x_k^b)\|_2^2
\end{eqnarray*}
Define $\kappa$ the maximum of the spectral norm of the four matrices, i.e.
\begin{eqnarray}
    \kappa = \max\left(\|M^1\|_2, \|M^2\|_2, \|M^3\|_2, \|M^4\|_2 \right) \label{eqn:kappa}
\end{eqnarray}
where $\|\cdot\|_2$ stands for the spectral norm of matrices.
\begin{thm} \label{thm:dual-recovery}
Assume $\ell(z)$ is $\gamma$-smooth. Let ${\bm \alpha}_*$ be the optimal solution to the dual problem in (\ref{eqn:dual}), and let $\ah_*$ be the approximately optimal solution for (\ref{eqn:dual2}) with suboptimality $\eta$. Then, with a probability at least $1 - \delta$, we have
\[
    \|{\bm \alpha}_* - \ah_* \|_2 \leq \max\left(8\epsilon\gamma\kappa\|\as\|_2, \sqrt{2\gamma\eta}\right)
\]
where $\kappa$ is define in (\ref{eqn:kappa}), provided $m \geq \frac{8}{\epsilon^2}\ln\frac{8N}{\delta}$.
\end{thm}
The proof of Theorem~\ref{thm:dual-recovery} can be found in the appendix. Unlike Theorem~\ref{thm:low-rank} where the data matrix $X$ is assumed to be low rank, Theorem~\ref{thm:dual-recovery} holds without any prior assumption about the data matrix. It shows that despite the random projection, the dual solution can be recovered approximately using the randomly projected vectors, provided that the number of random projections $m$ is sufficiently large, $\kappa$ is small, and the approximately optimal solution $\ah_*$ is sufficiently accurate. In the case when most of the training examples are not linear dependent, we could have $\kappa = \Theta(N/d)$, which could be a modest number when $d$ is very large. The result in Theorem~\ref{thm:dual-recovery} essentially justifies the key idea of our approach, i.e. computing the dual variables first and recovering the distance metric later. Finally, since $\|{\bm \alpha}_* - \ah_* \|_2$, the approximation error in the recovered dual variables, is proportional to the square root of the suboptimality $\eta$, an accurate solution for (\ref{eqn:dual2}) is needed to ensure a small approximation error. We note that given Theorem~\ref{thm:dual-recovery}, it is straightforward to bound $\|M_* - \M\|_2$ using the relationship between the dual variables and the primal variables in (\ref{eqn:primal-dual}).

\section{Experiments}
\label{sec:exp}

We will first describe the experimental setting, and then present our empirical study for ranking and classification tasks on various datesets.

\subsection{Experimental Setting}

\begin{table}[!ht]
\centering
\caption{Statistics for the datasets used in our empirical study. \#C is the number of classes. \#F is the number of original features. \#Train and \#Test represent the number of training data and test data, respectively.}
\label{rlr}
\begin{tabular}{|c|c|c|c|c|}
\hline
          &\# C &\# F&\#Train &\#Test\\\hline
{\it usps}      &10       &256         &7,291    &2,007       \\\hline
{\it protein}   &3        &357         &17,766   &6,621       \\\hline
{\it caltech30} &30       &1,000       &5,502    &2,355       \\\hline
{\it tdt30}     &30       &1,000       &6,575    &2,819      \\\hline
{\it 20news}    &20       &1,000       &15,935   &3,993      \\\hline
{\it rcv30}     &30       &1,000       &507,585  &15,195     \\\hline
\end{tabular}
\end{table}


\paragraph{Data sets} Six datasets are used to validate the effectiveness of the proposed algorithm for DML. Table~\ref{rlr} summarizes the information of these datasets. {\it caltech30} is a subset of Caltech256 image dataset~\citep{Griffin2007} and we use the version pre-processed by~\citep{ChechikSSB10}. {\it tdt30} is a subset of tdt2 dataset~\citep{CaiWH09}. Both {\it caltech30} and {\it tdt30} are comprised of the examples from the $30$ most popular categories. All the other datasets are downloaded from LIBSVM~\citep{ChangL11}, where {\it rcv30} is a subset of the original dataset consisted of documents from the $30$ most popular categories. For datasets {\it tdt30}, {\it 20news} and {\it rcv30}, they are comprised of documents represented by vectors of $\sim 50,000$ dimensions. Since it is expensive to compute and maintain a matrix of $50,000\times50,000$, for these three datasets, we follow the procedure in~\citep{ChechikSSB10} that maps all documents to a space of $1,000$ dimension. More specifically, we first keep the top $20,000$ most popular words for each collection, and then reduce their dimensionality to $1,000$ by using PCA. We emphasize that for several data sets in our test beds, their data matrices can not be well approximated by low rank matrices. Fig.~\ref{rank} summarizes the eigenvalue distribution of the six datasets used in our experiment. We observe that four out of these datasets (i.e., {\it caltech20}, {\it tdt30}, {\it 20news}, {\it rcv30}) have a flat eigenvalue distribution, indicating that the associated data matrices can not be well approximated by a low rank matrix. This justifies the importance of removing the low rank assumption from the theory of dual random projection, an important contribution of this work.

\begin{figure*}[!ht]
\centering
\begin{minipage}[h]{1.9in}
\centering
\includegraphics[width= 2.0in ]{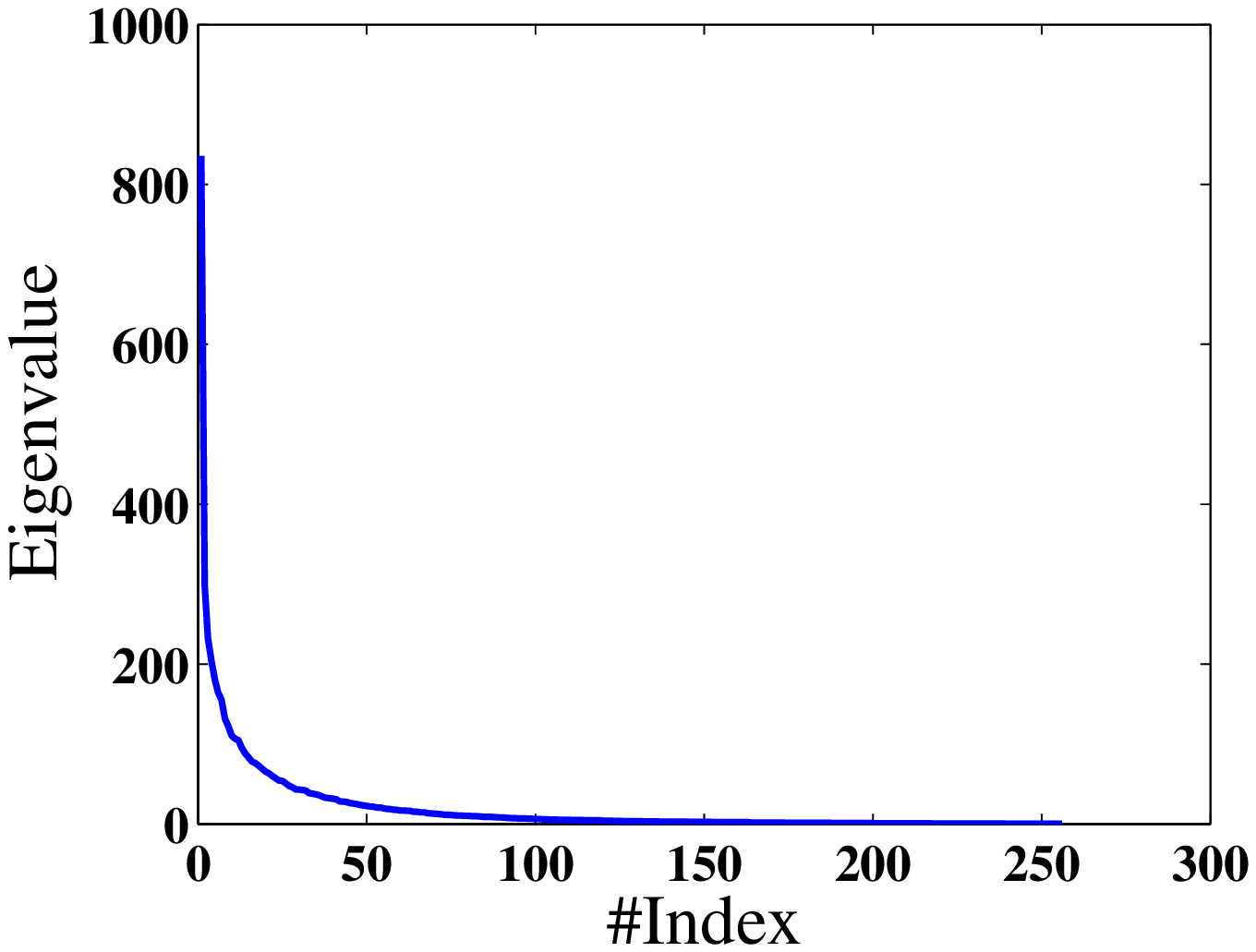}\\
\mbox{\footnotesize (a) {\it usps}}
\end{minipage}
\begin{minipage}[h]{1.9in}
\centering
\includegraphics[width= 2.0in ]{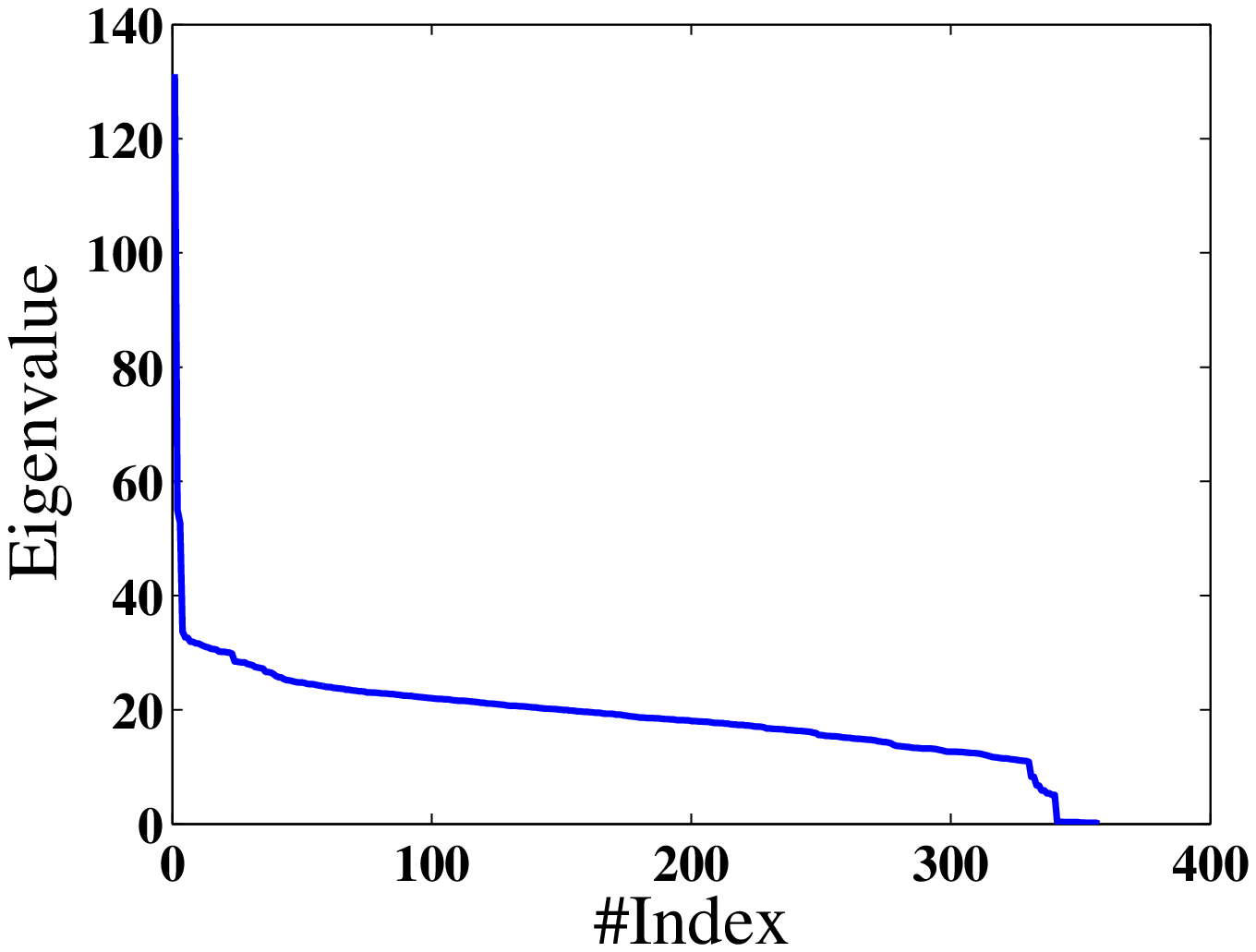}\\
\mbox{\footnotesize (b) {\it protein}}
\end{minipage}
\begin{minipage}[h]{1.9in}
\centering
\includegraphics[width= 2.0in ]{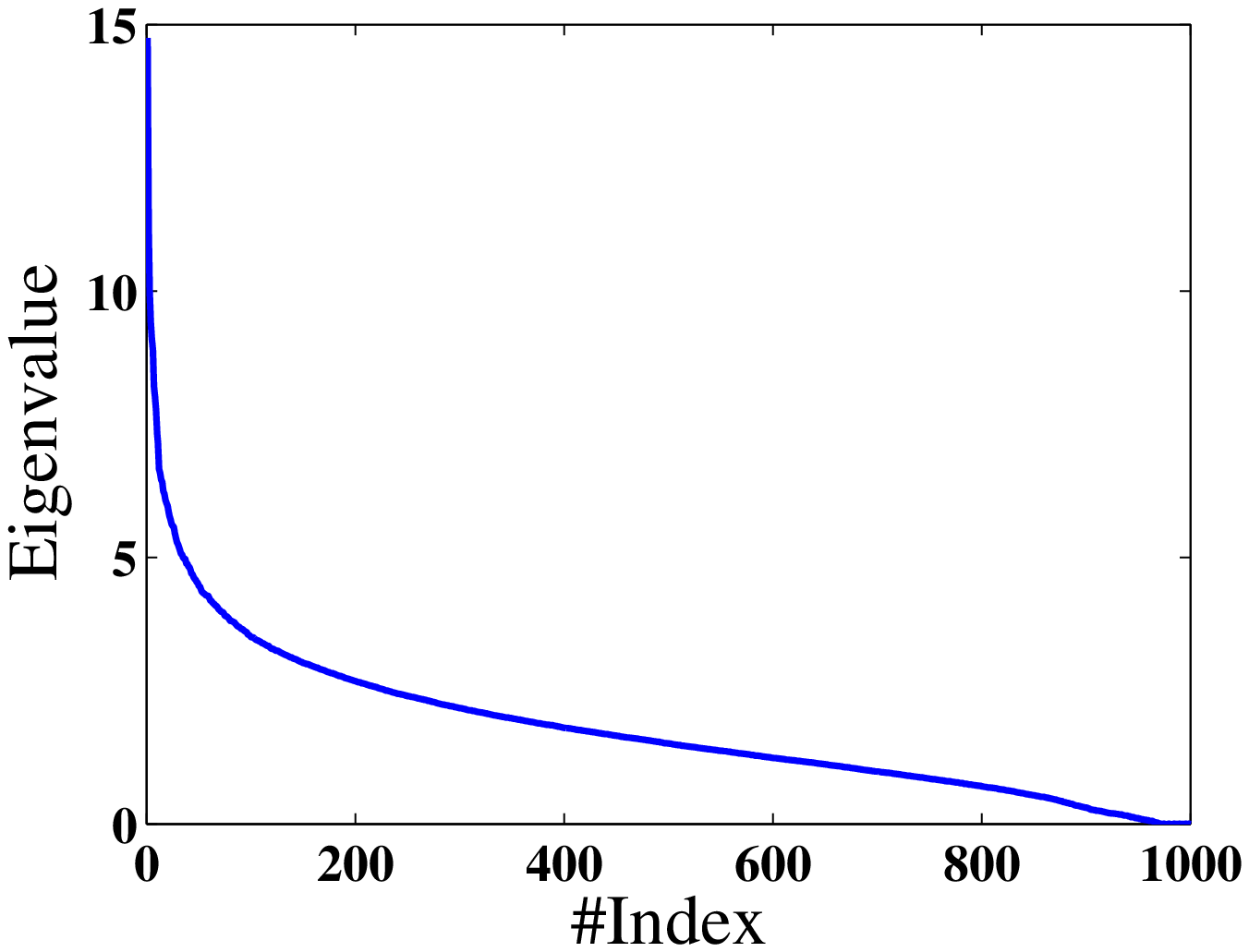}\\
\mbox{\footnotesize (c) {\it caltech30}}
\end{minipage}

\begin{minipage}[h]{1.9in}
\centering
\includegraphics[width= 2.0in ]{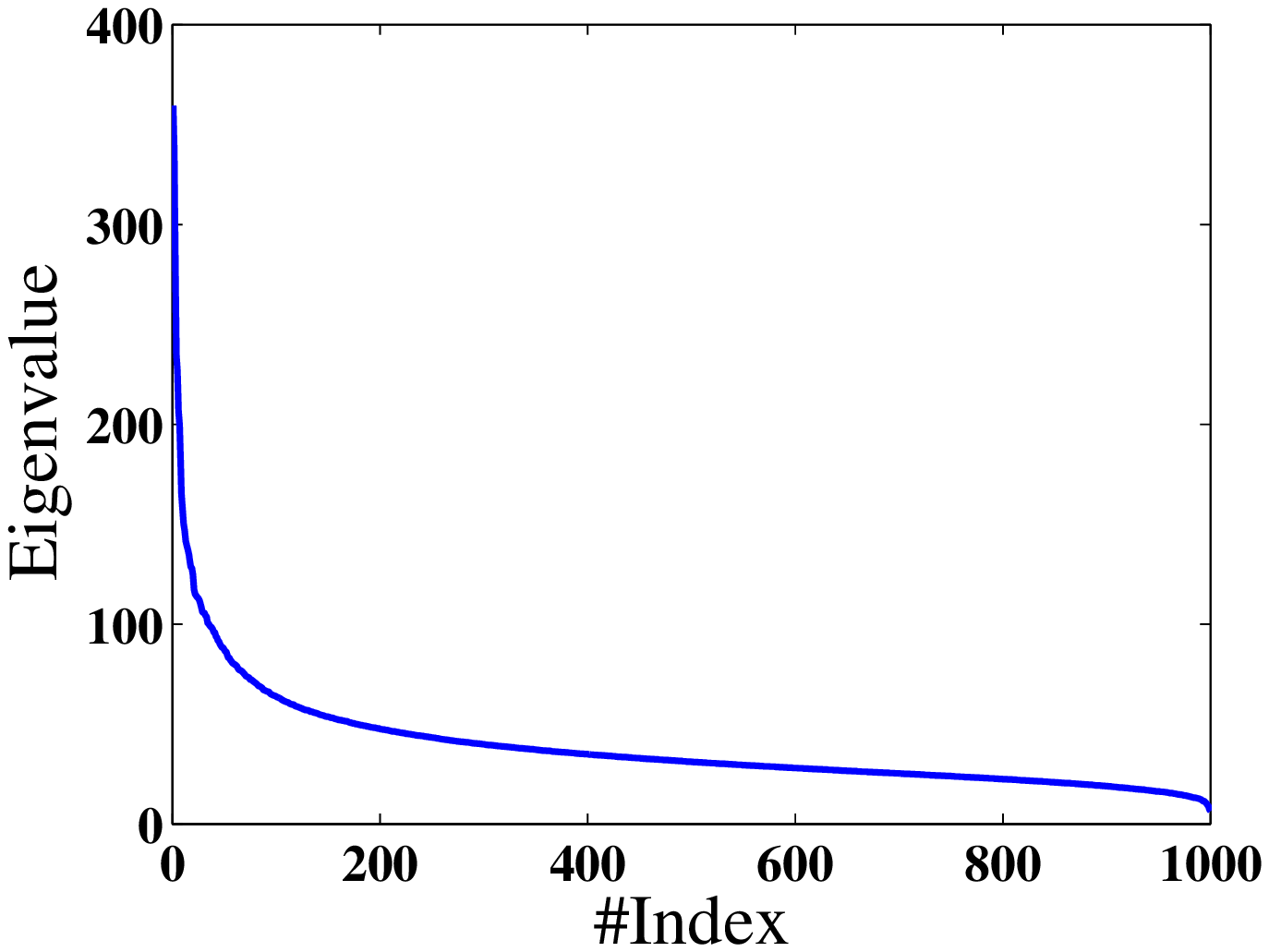}\\
\mbox{\footnotesize (d) {\it tdt30}}
\end{minipage}
\begin{minipage}[h]{1.9in}
\centering
\includegraphics[width= 2.0in ]{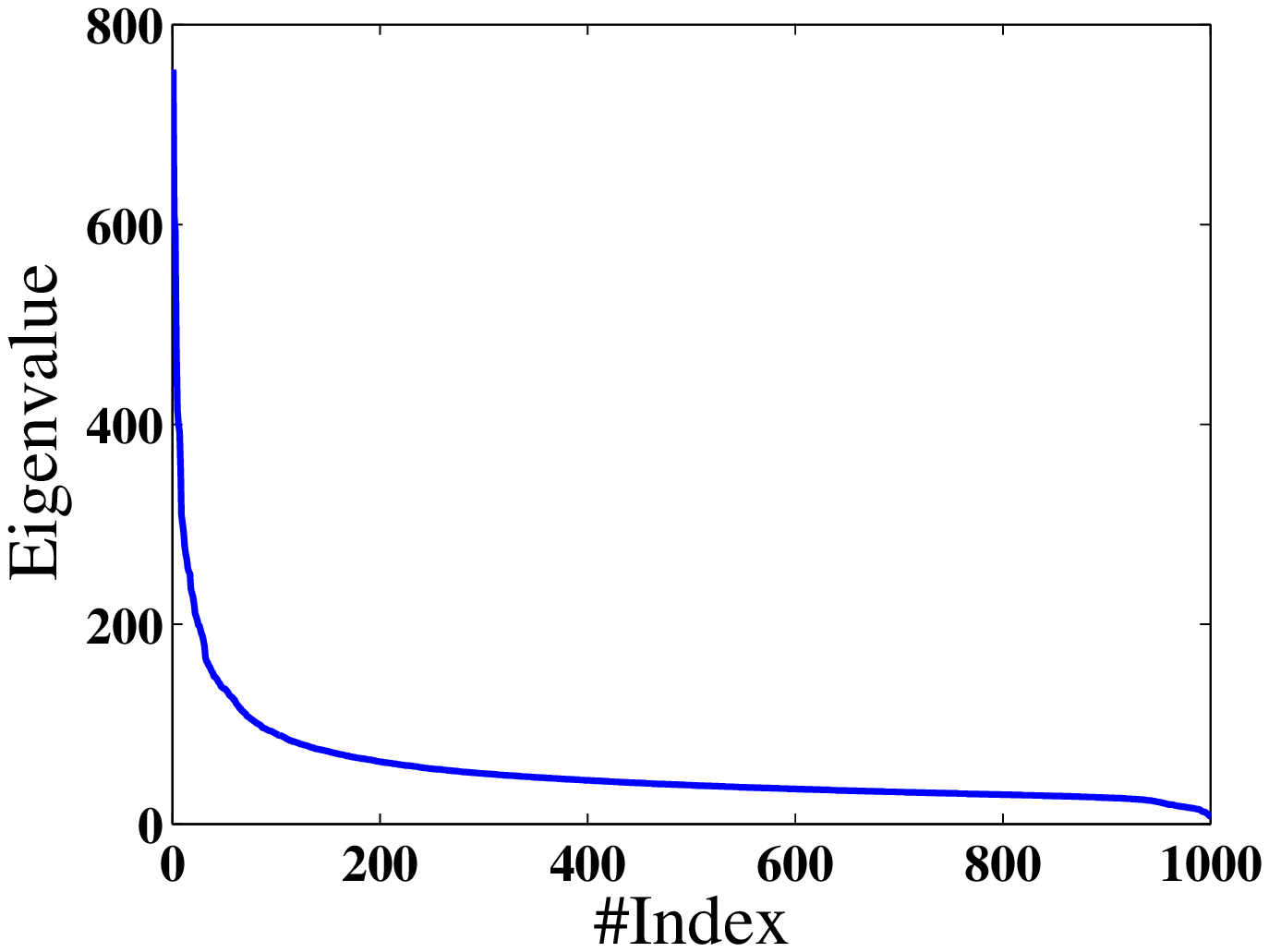}\\
\mbox{\footnotesize (e) {\it 20news}}
\end{minipage}
\begin{minipage}[h]{1.9in}
\centering
\includegraphics[width= 2.0in ]{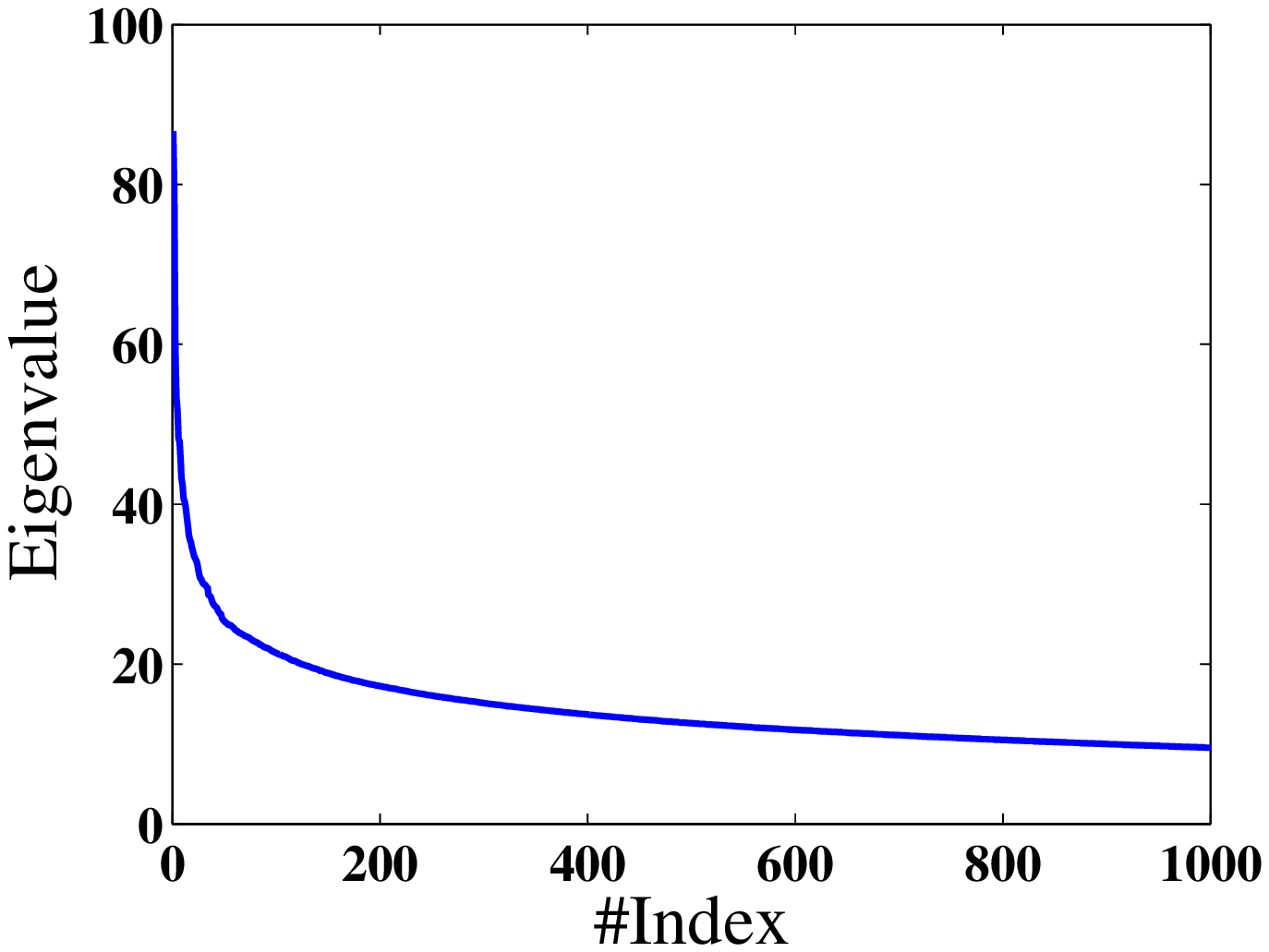}\\
\mbox{\footnotesize (f) {\it rcv30}}
\end{minipage}
\caption{The eigenvalue distribution of datasets used in our empirical study}\label{rank}
\end{figure*}

For most datasets used in this study, we use the standard training/testing split provided by the original datasets, except for datasets {\it tdt30}, {\it caltech30} and {\it rcv30}. For {\it tdt30} and {\it caltech30}, we randomly select $70\%$ of the data for training and use the remaining $30\%$ for testing; for {\it rcv30}, we switch the training and test sets defined by the original package to ensure that the number of training examples is sufficiently large.

\paragraph{Evaluation metrics} To measure the quality of learned distance metrics, two types of evaluations are adopted in our study. First, we follow the evaluation protocol in~\citep{ChechikSSB10} and  evaluate the learned metric by its ranking performance. More specifically, we treat each test instance $\q$ as a query, and rank the other test instances in the ascending order of their distance to $\q$ using the learned metric. The mean-average-precision(mAP) given below is used to evaluate the quality of the ranking list
\vspace{-3mm}
\begin{eqnarray*}
mAP = \frac{1}{|Q|}\sum_{i=1}^{|Q|}\frac{1}{r_i}\sum_{j=1}^{r_i}P(\x_{i\sim j})
\end{eqnarray*}
where $|Q|$ is the size of query set, $r_i$ is the number of relevant instances for $i$-th query and $P(\x_{i\sim j})$ is the precision for the first $j$ ranked instances when the instance ranked at the $j$-th position is relevant to the query $\q$. Here, an instance $\x$ is relevant to a query $\q$ if they belong to the same class. Second, we evaluate the learned metric by its classification performance with $k$-nearest neighbor classifier. More specifically, for each test instance $\q$, we apply the learned metric to find the first $k$ training examples with the shortest distance, and predict the class assignment for $k$ by taking the majority vote among the $k$ nearest neighbors. Finally, we also evaluate the computational efficiency of the proposed algorithm for DML by its efficiency.

\paragraph{Baselines} Besides the Euclidean distance that is used as a baseline similarity measure, six state-of-the-art DML methods are compared in our empirical study:
\begin{compactitem}
\item {\bf DuOri}: This algorithm first applies Combined Stochastic Dual Coordinate Ascent (CSDCA)~\citep{tzhang2012} to solve the dual problem in (\ref{eqn:dual}) and then computes the distance metric using the learned dual variables.
\item {\bf DuRP}: This is the proposed algorithm for DML (i.e. Algorithm~\ref{alg:1}).
\item {\bf SRP}: This algorithm applies random projection to project data into low dimensional space, and then it employs CSDCA to learn the distance metric in this subspace.
\item {\bf SPCA}: This algorithm uses PCA as the initial step to reduce the dimensionality, and then applies CSDCA to learn the distance metric in the subspace generated by PCA.
\item {\bf OASIS}~\citep{ChechikSSB10}: A state-of-art online learning algorithm for DML that learns the optimal distance metric directly from the original space without any dimensionality reduction.
\item {\bf LMNN}~\citep{weinberger2009}: A state-of-art batch learning algorithm for DML. It performs the dimensionality reduction using PCA before starting DML.
\end{compactitem}

\paragraph{Implementation details} We randomly select $N = 100,000$ active triplets (i.e., incur the positive hinge loss by Euclidean distance) and set the number of epochs to be $3$ for all stochastic methods (i.e., DuOri, DuRP, SRP, SPCA and OASIS), which yields sufficiently accurate solutions in our experiments and is also consistent with the observation in \citep{tzhang2012}. We search $\lambda$ in $\{10^{-5}, 10^{-4}, 10^{-3}, 10^{-2}\}$ and fix it as $1/N$ since it is insensitive. The step size of CSDCA is set according to the analysis in \citep{tzhang2012}. For all stochastic optimization methods, we follow the one-projection paradigm by projecting the learned metric onto the PSD cone. The hinge loss is used in the implementation of the proposed algorithm. Both OASIS and LMNN use the implementation provided by the original authors and parameters are tuned based on the recommendation by the original authors. All methods are implemented in Matlab, except for LMNN, whose core part is implemented in C, which is shown to be more efficient than our Matlab implementation. All stochastic optimization methods are repeated five times and the average result over five trials is reported. All experiments are implemented on a Linux Server with 64GB memory and $12\times 2.4$GHz CPUs and only single thread is permitted for each experiment.

\subsection{Efficiency of the Proposed Method}

\begin{table*}[!ht]
\centering
\caption{CPUtime (minutes) for different methods for DML. All algorithms are implemented in Matlab except for LMNN whose core part is implemented in C and is more efficient than our Matlab implementation.}
\label{rlr3}
\begin{tabular}{|c|c|c|c|c|c|c|}
\hline
         &\multicolumn{3}{c|}{Metric in Original Space}&\multicolumn{3}{c|}{Metric in Subspace}\\\hline
         &DuOri                &DuRP           &OASIS             & SRP           &SPCA           &LMNN\\\hline
{\it usps}     &77.0$\pm$3.2      &0.3$\pm$0.0 &6.2$\pm$0.2       &0.2$\pm$0.0&0.2$\pm$0.0&14.2\\\hline
{\it protein}  &214.5$\pm$5.8     &0.6$\pm$0.0 &81.9$\pm$3.3      &0.2$\pm$0.0&0.2$\pm$0.0&488.9\\\hline
{\it caltech30}&1,214.5$\pm$229.5 &1.3$\pm$0.0 &640.4$\pm$121.2   &0.2$\pm$0.0&0.5$\pm$0.0&2,197.9\\\hline
{\it tdt30}    &1,029.9$\pm$16.8  &0.8$\pm$0.0 &140.8$\pm$4.5     &0.2$\pm$0.0&0.4$\pm$0.0&624.2\\\hline
{\it 20news}   &1,212.9$\pm$154.3 &1.0$\pm$0.0 &216.3$\pm$48.8    &0.2$\pm$0.0&0.5$\pm$0.0&1,893.6\\\hline
{\it rcv30}    &1,121.3$\pm$79.4  &1.3$\pm$0.0 &432.5$\pm$7.7     &0.2$\pm$0.0&4.2$\pm$0.0&N/A\\\hline
\end{tabular}
\end{table*}

In this experiment, we set the number of random projection to be $10$, which according to experimental results in Section~\ref{sec:exp:rank} and \ref{sec:exp:class}, yields almost the optimal performance for the proposed algorithm. For fair comparison, the number of reduced dimension is also set to be $10$ for LMNN.

Table.~\ref{rlr3} compares the CPUtime (in minutes) of different methods. Notice that the time of sampling triplets is not taken into account as it is consumed by all the methods, and all the other operators (e.g., random projection and PCA) are included. It is not surprising to observe that DuRP, SRP and SPCA have similar CPUtimes, and are significantly more efficient than the other methods due to the effect of dimensionality reduction. Since DuRP and SRP share the same procedure for computing the dual variables in the subspace, the only difference between them lies in the procedure for reconstructing the distance metric from the estimated dual variables, a computational overhead that makes DuRP slightly slower than SRP. For all datasets, we observe that DuRP is at least 200 times faster than DuOri and 20 times faster than OASIS. Compared to the stochastic optimization methods, LMNN is the least efficient on three datasets (i.e., {\it protein}, {\it caltech30} and {\it 20news}), mostly due to the fact that it is a batch learning algorithm.

\subsection{Evaluation by Ranking}
\label{sec:exp:rank}
\begin{table*}[!ht]
\centering
\caption{Comparison of ranking results measured by mAP ($\%$) for different metric learning algorithms.}
\label{rlr2}
\begin{tabular}{|c|c|c|c|c|c|c|c|}
\hline
       &\multicolumn{4}{c|}{Metric in Original Space}&\multicolumn{3}{c|}{Metric in Subspace Metric}\\\hline
       &Euclid &DuOri        &DuRP      &OASIS  &SRP &SPCA&LMNN    \\\hline
{\it usps}      &53.6   &67.7$\pm$1.7 &67.1$\pm$1.2   &62.5$\pm$0.5  &32.6$\pm$5.4 &41.6$\pm$0.4 &59.8\\\hline
{\it protein}   &39.0   &47.0$\pm$0.1 &49.1$\pm$0.1   &45.7$\pm$0.1  &37.7$\pm$0.1 &41.9$\pm$0.1 &41.9\\\hline
{\it caltech30} &16.4   &23.8$\pm$0.1 &25.5$\pm$0.1   &25.4$\pm$0.2  &8.1$\pm$0.4  &19.5$\pm$0.0 &16.3\\\hline
{\it tdt30}     &36.8   &65.9$\pm$0.2 &69.4$\pm$0.3   &55.9$\pm$0.1  &11.2$\pm$0.3 &49.7$\pm$0.2 &66.4\\\hline
{\it 20news}    &8.4    &20.1$\pm$0.2 &24.9$\pm$0.3   &16.2$\pm$0.1  &5.3$\pm$0.1  &12.2$\pm$0.1 &22.5\\\hline
{\it rcv30}     &16.7   &65.7$\pm$0.1 &63.2$\pm$0.2   &68.6$\pm$0.1  &12.8$\pm$0.4 &46.5$\pm$0.0 &N/A\\\hline
\end{tabular}
\end{table*}

In first experiment, we set the number of random projections used by SRP, SPCA and the proposed DuRP algorithm to be $10$, which is roughly $1\%$ of the dimensionality of the original space. For fair comparison, the number of reduced dimension for LMNN is also set to be $10$. We measure the quality of learned metrics by its ranking performance using the metric of mAP.

Table.~\ref{rlr2} summarizes the performance of different methods for DML. First, we observe that DuRP significantly outperforms SRP and SPCA for all datasets. In fact, SRP is worse than Euclidean distance which computes the distance in the original space. SPCA is only able to perform better than the Euclidean distance, and is outperformed by all the other DML algorithms. Second, we observe that for all the datasets, DuRP yields similar performance as DuOri. The only difference between DuRP and DuOri is that DuOri solves the dual problem without using random projection. The comparison between DuRP and DuOri indicates that the random projection step has minimal impact on the learned distance metric, justifying the design of the proposed algorithm. Third, compared to OASIS, we observe that DuRP performs significantly better on two datasets (i.e., {\it tdt30} and {\it 20news}) and has the comparable performance on the other datasets. Finally, we observe that for all datasets, the proposed DuRP method significantly outperforms LMNN, a state-of-the-art batch learning algorithm for DML. We also note that because of limited memory, we are unable to run LMNN on datasets {\it rcv30}.

In the second experiment, we vary the number of random projections from $10$ to $50$. All stochastic methods are run with five trails and Fig.~\ref{rank} reports the average results with standard deviation. Note that the performance of OASIS and DuOri remain unchanged with varied number of projections because they do not use projection. It is surprising to observe that DuRP almost achieves its best performance with only 10 projections for all datasets. This is in contrast to SRP and SPCA, whose performance usually improves with increasing number of projections except for the data set {\it usps} where the performance of SPCA declines when the number of random projections is increased from $10$ to $30$. A detailed examination shows that the strange behavior for SPCA is due to its extreme low rank at 30 projections after the learned matrix is projected onto the PSD cone. More investigation is needed for this strange case. We also observe that DuRP outperforms DuOri for several datasets (i.e. {\it protein}, {\it caltech30}, {\it tdt30} and {\it 20news}). We suspect that the better performance of DuRP is because of the implicit regularization due to the random projection. We plan to investigate more about the regularization capability of random projection in the future. We finally point out that with sufficiently large number of projections, SPCA is able to outperform OASIS on 3 datasets (i.e., {\it protein}, {\it tdt30} and {\it 20news}), indicating that the comparison result may be sensitive to the number of projections.

\begin{figure*}[!ht]
\centering
\begin{minipage}[h]{1.9in}
\centering
\includegraphics[width= 2.0in ]{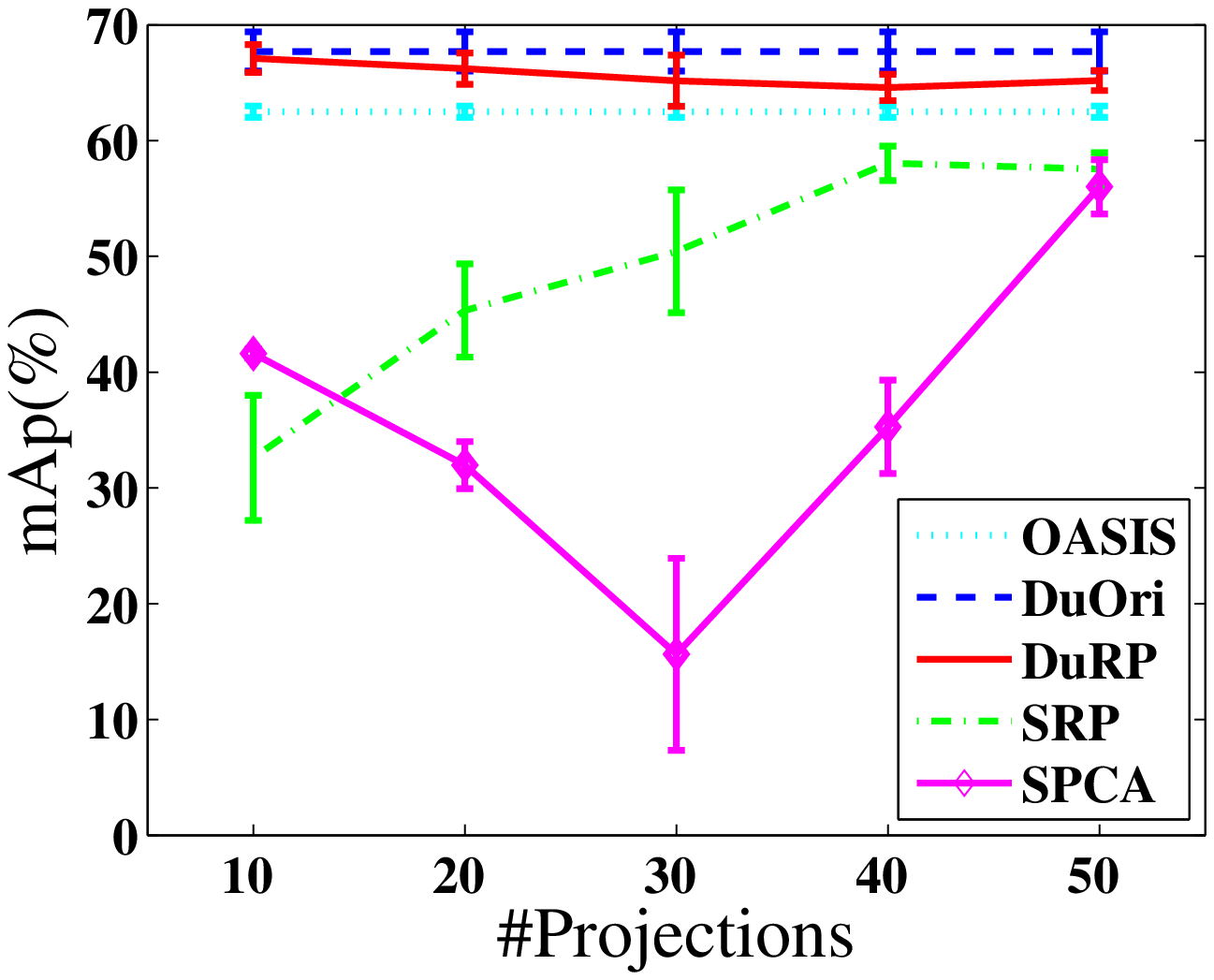}\\
\mbox{\footnotesize (a) {\it usps}}
\end{minipage}
\begin{minipage}[h]{1.9in}
\centering
\includegraphics[width= 2.0in ]{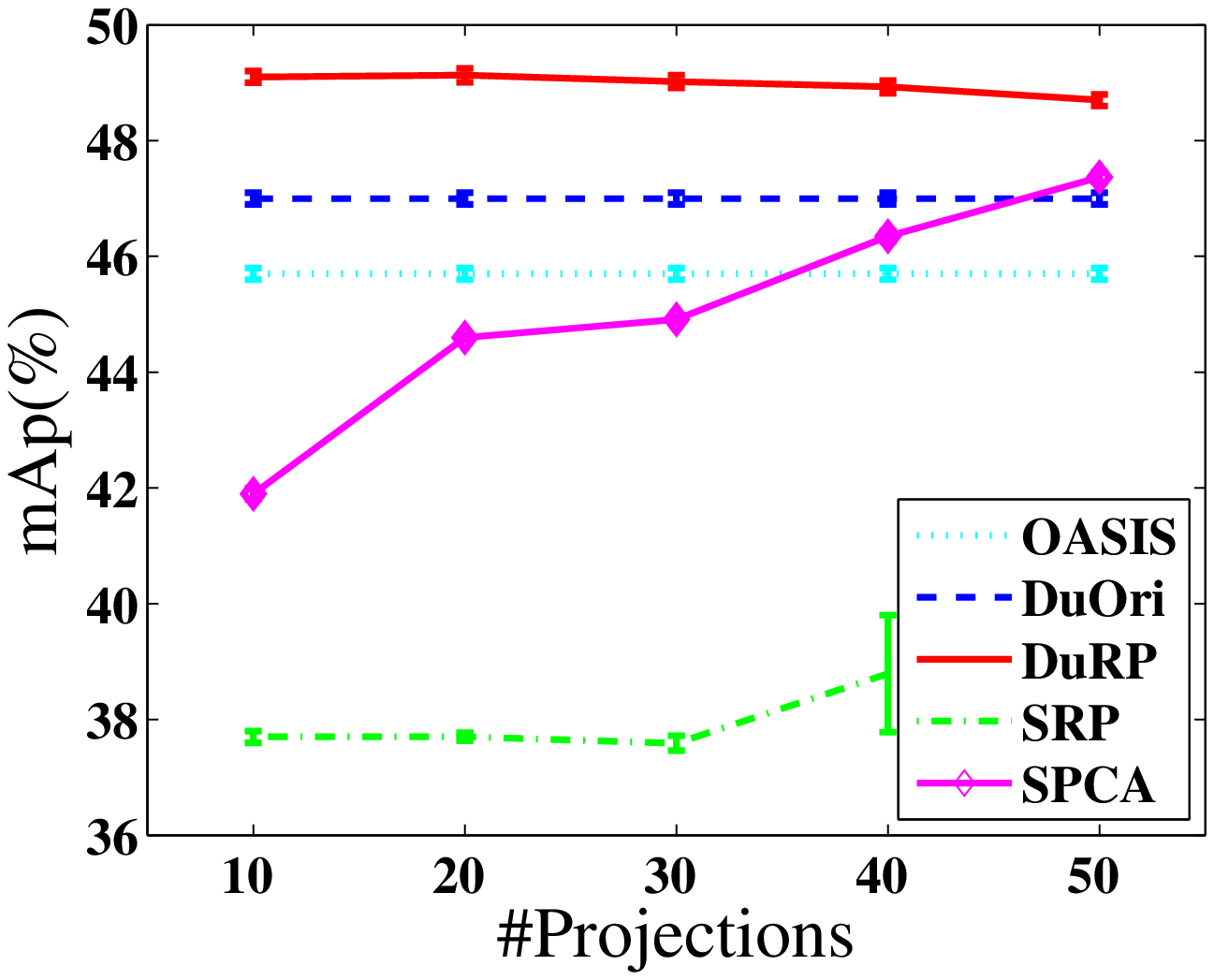}\\
\mbox{\footnotesize (b) {\it protein}}
\end{minipage}
\begin{minipage}[h]{1.9in}
\centering
\includegraphics[width= 2.0in ]{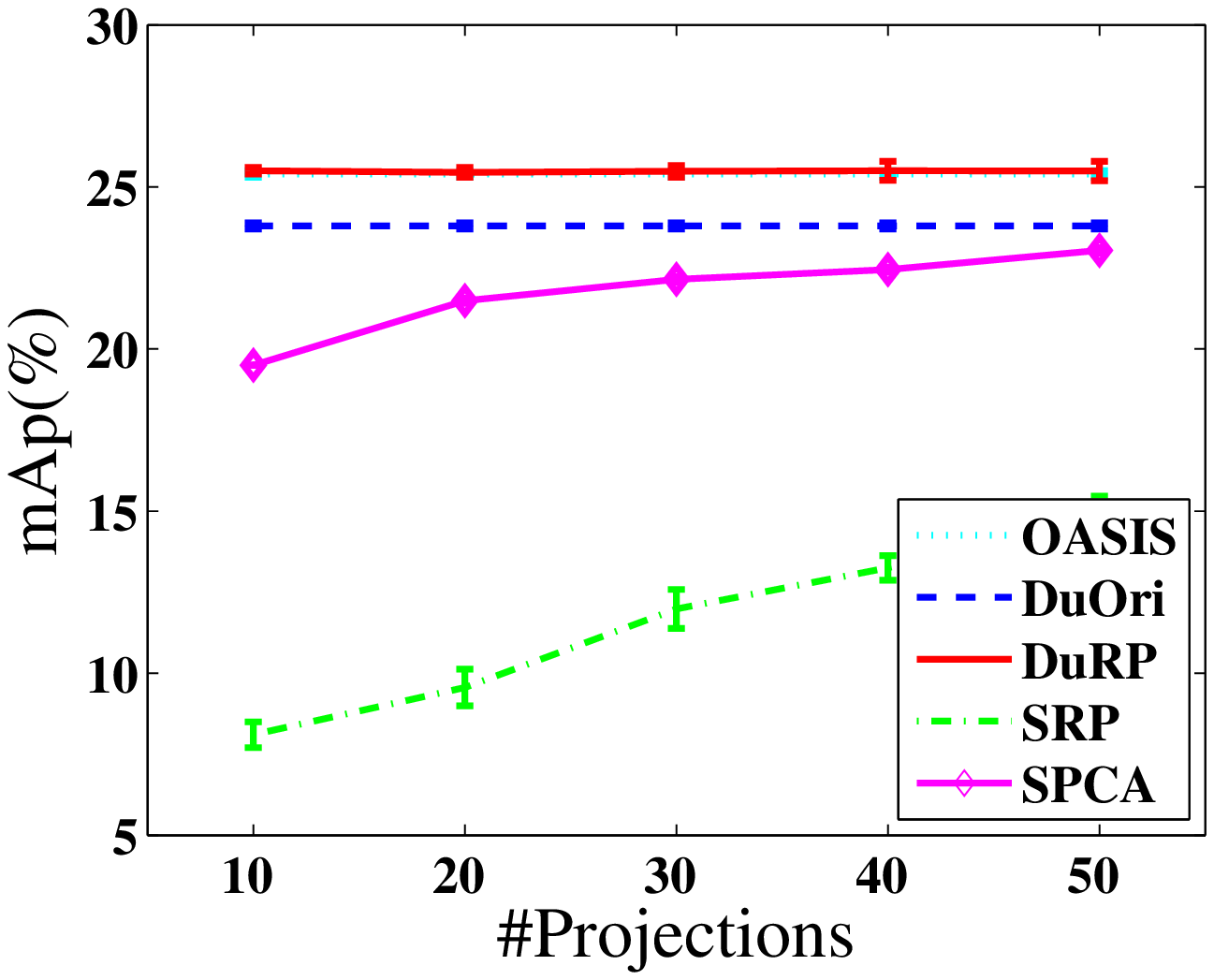}\\
\mbox{\footnotesize (c) {\it caltech30}}
\end{minipage}

\begin{minipage}[h]{1.9in}
\centering
\includegraphics[width= 2.0in ]{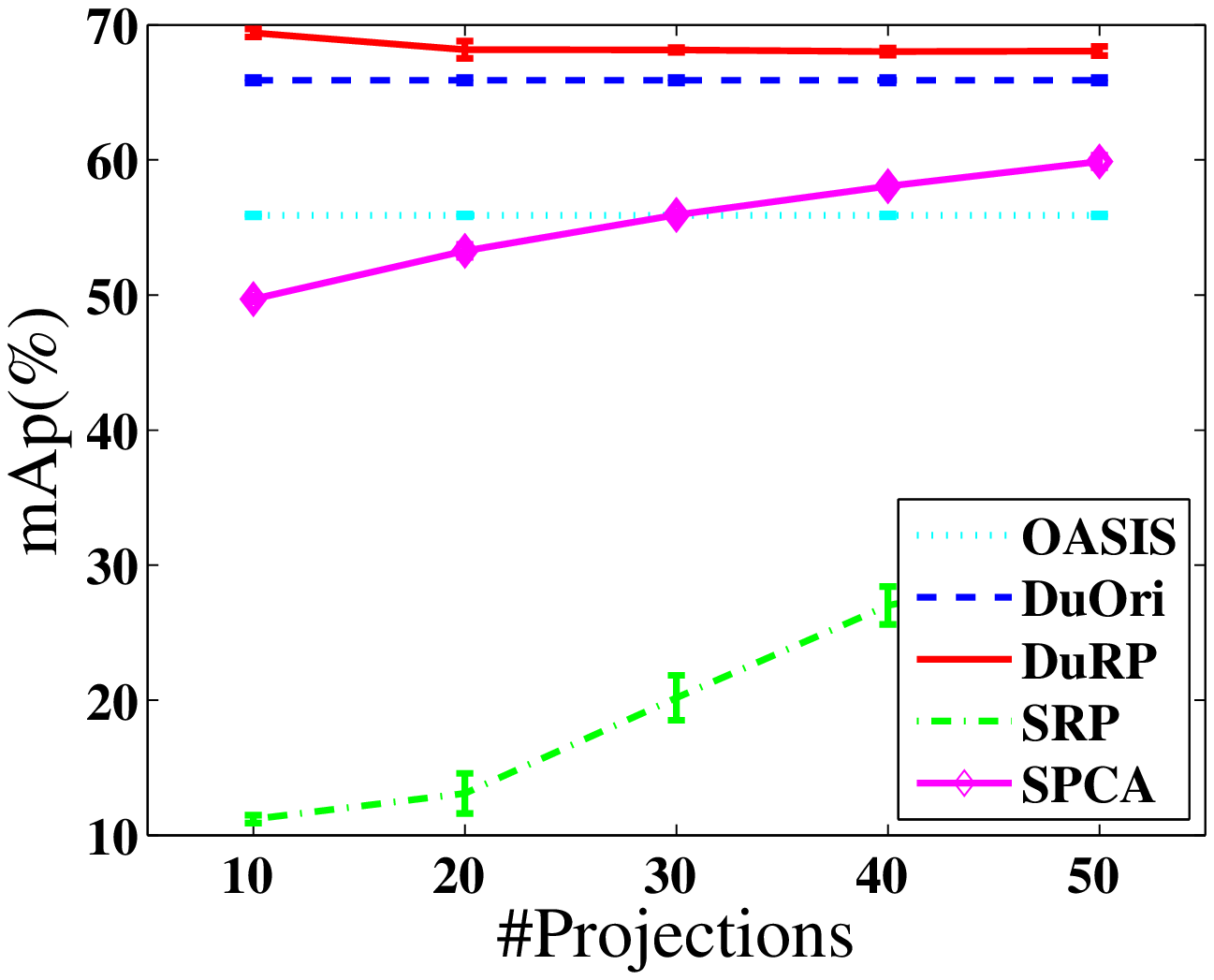}\\
\mbox{\footnotesize (d) {\it tdt30}}
\end{minipage}
\begin{minipage}[h]{1.9in}
\centering
\includegraphics[width= 2.0in ]{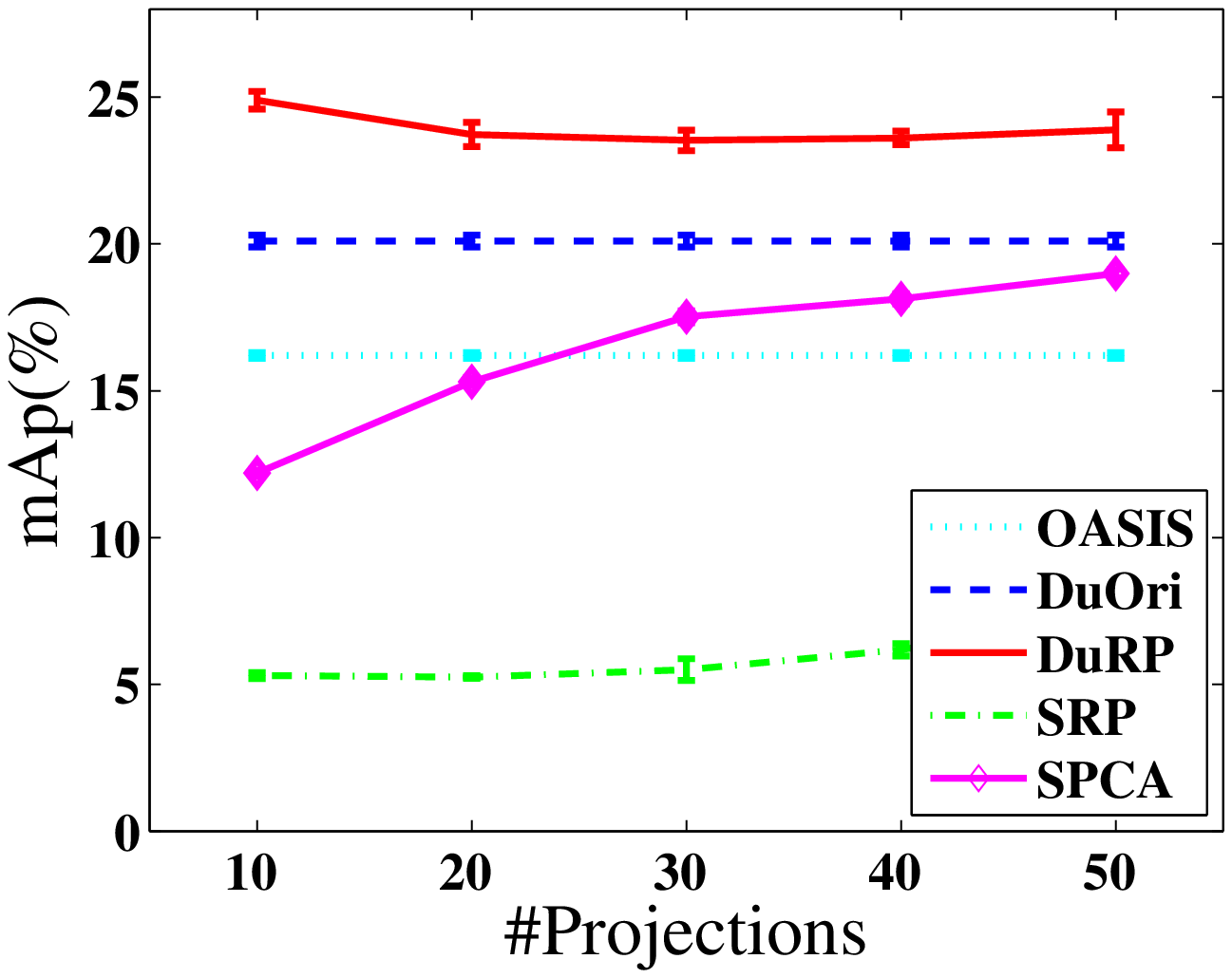}\\
\mbox{\footnotesize (e) {\it 20news}}
\end{minipage}
\begin{minipage}[h]{1.9in}
\centering
\includegraphics[width= 2.0in ]{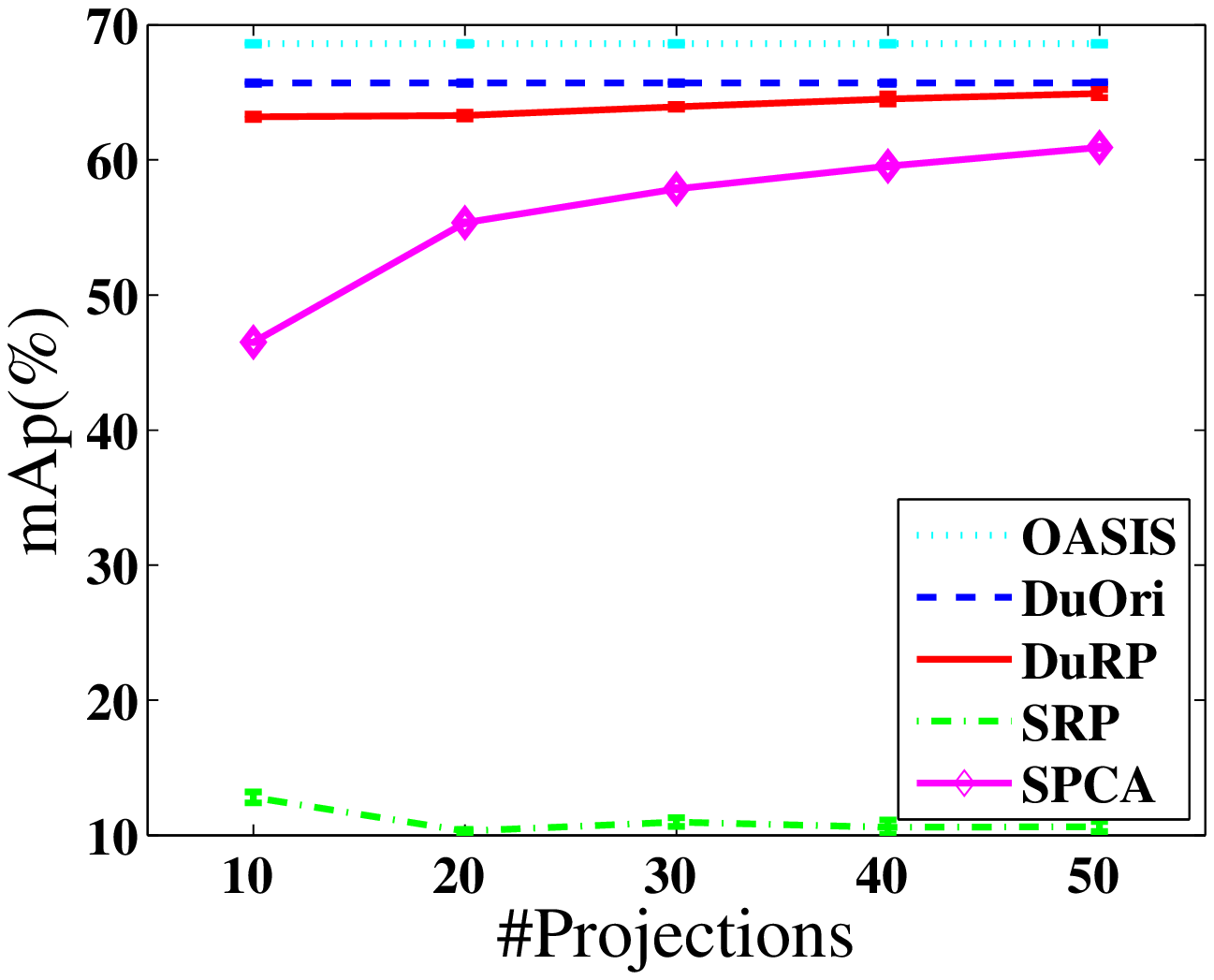}\\
\mbox{\footnotesize (f) {\it rcv30}}
\end{minipage}
\caption{The comparison of different stochastic algorithms for ranking}\label{rank}
\end{figure*}

\subsection{Evaluation by Classification}
\label{sec:exp:class}
\begin{figure*}[!ht]
\centering
\begin{minipage}[h]{1.9in}
\centering
\includegraphics[width= 2.0in ]{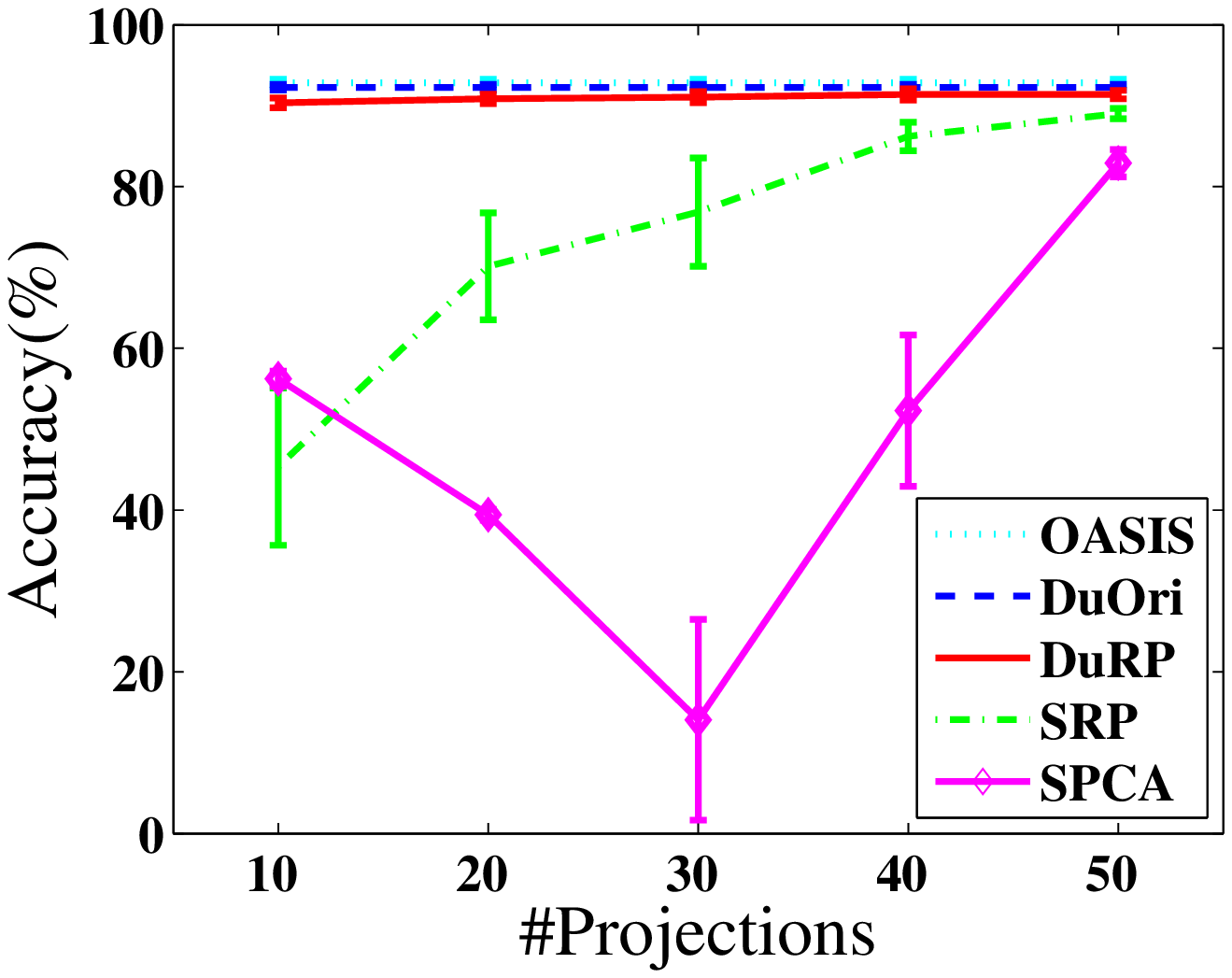}\\
\mbox{\footnotesize (a) {\it usps}}
\end{minipage}
\begin{minipage}[h]{1.9in}
\centering
\includegraphics[width= 2.0in ]{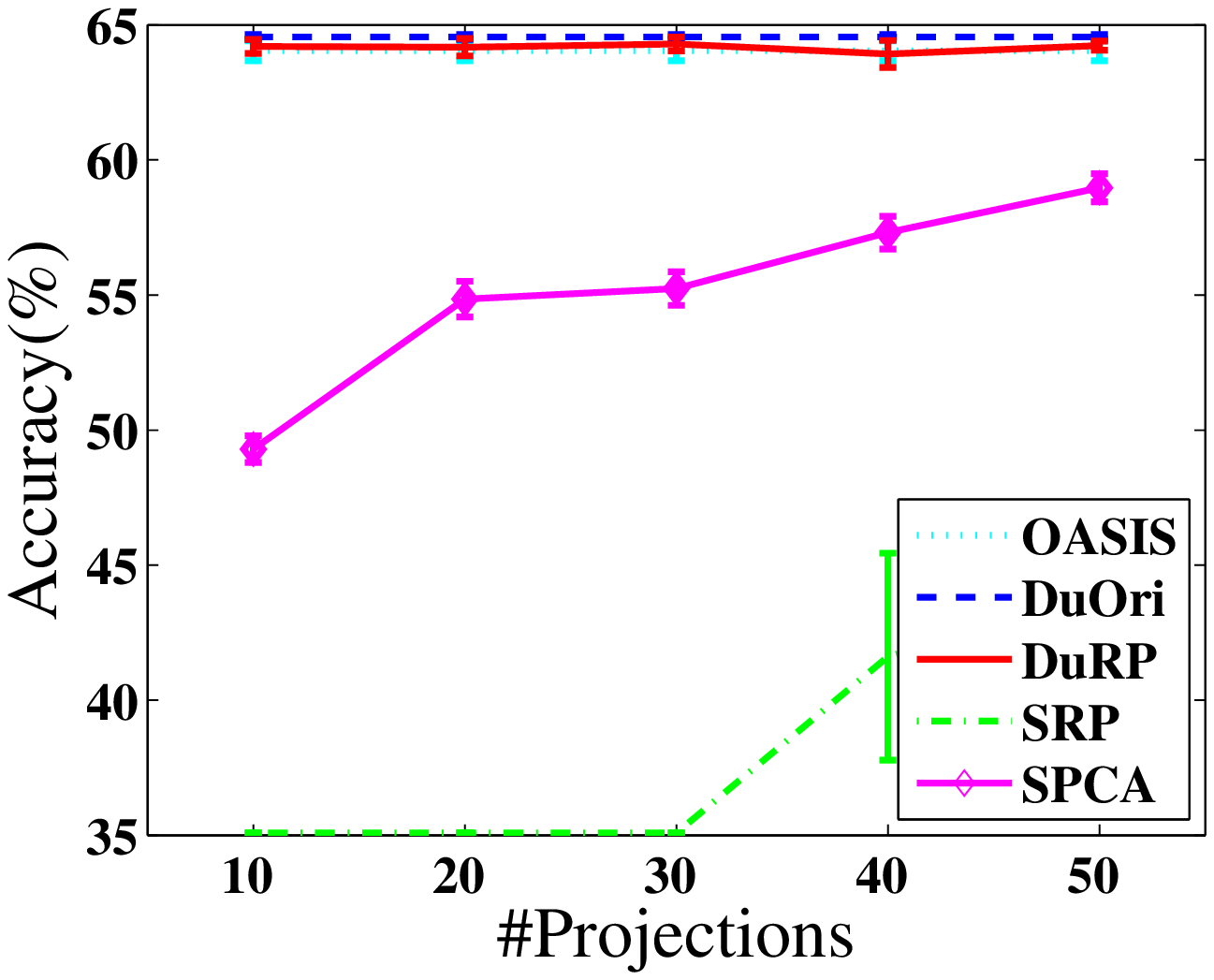}\\
\mbox{\footnotesize (b) {\it protein}}
\end{minipage}
\begin{minipage}[h]{1.9in}
\centering
\includegraphics[width= 2.0in ]{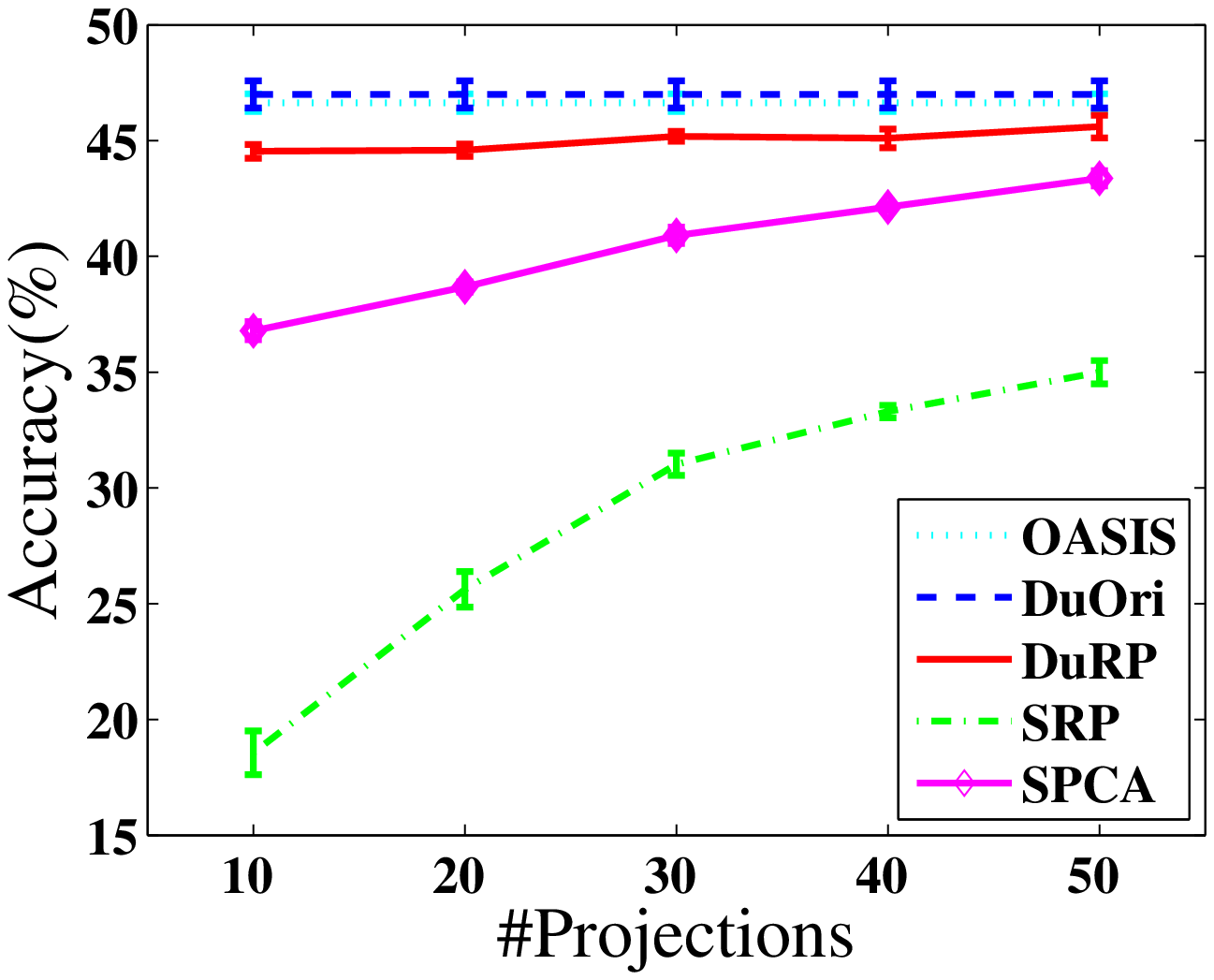}\\
\mbox{\footnotesize (c) {\it caltech30}}
\end{minipage}

\begin{minipage}[h]{1.9in}
\centering
\includegraphics[width= 2.0in ]{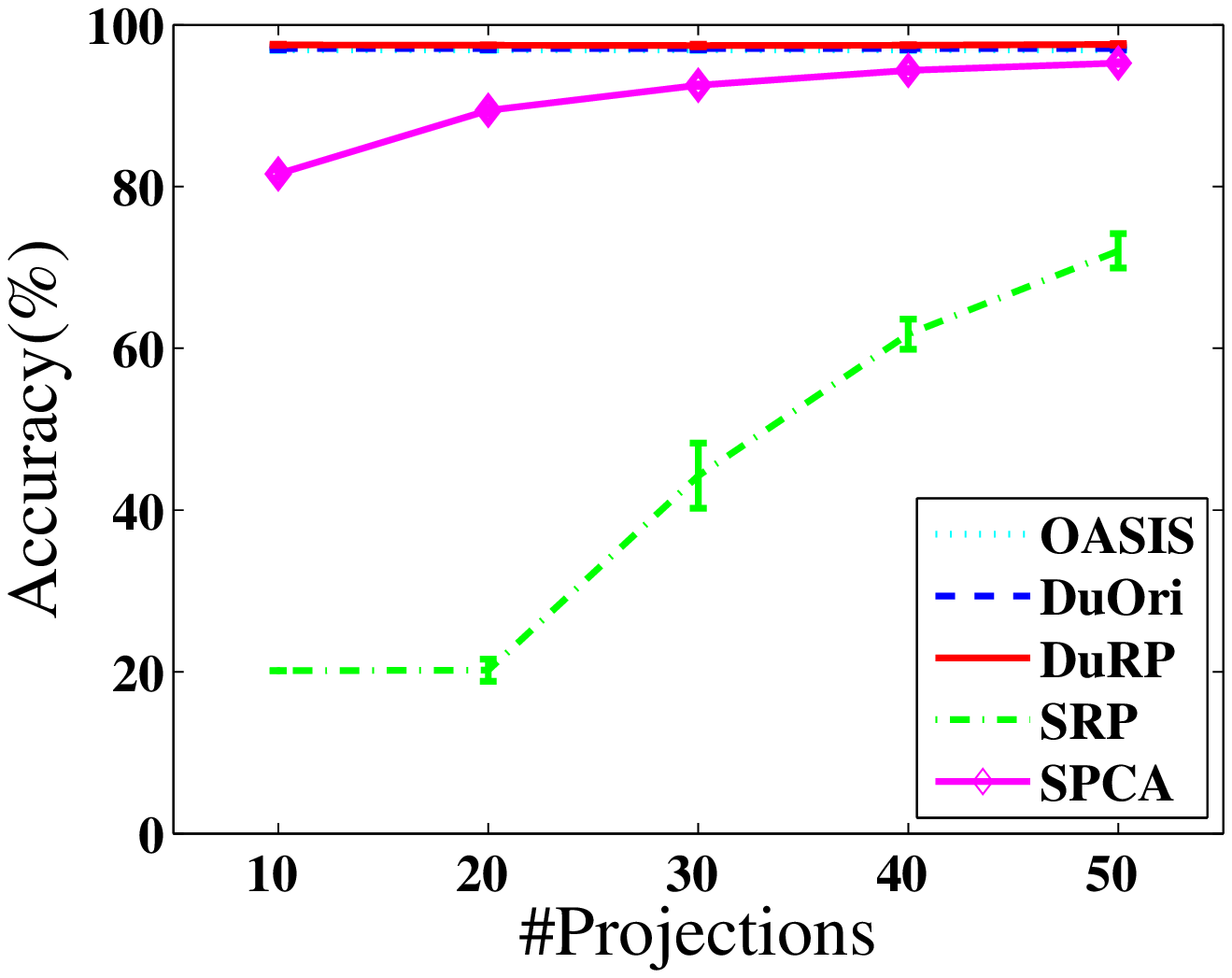}\\
\mbox{\footnotesize (d) {\it tdt30}}
\end{minipage}
\begin{minipage}[h]{1.9in}
\centering
\includegraphics[width= 2.0in ]{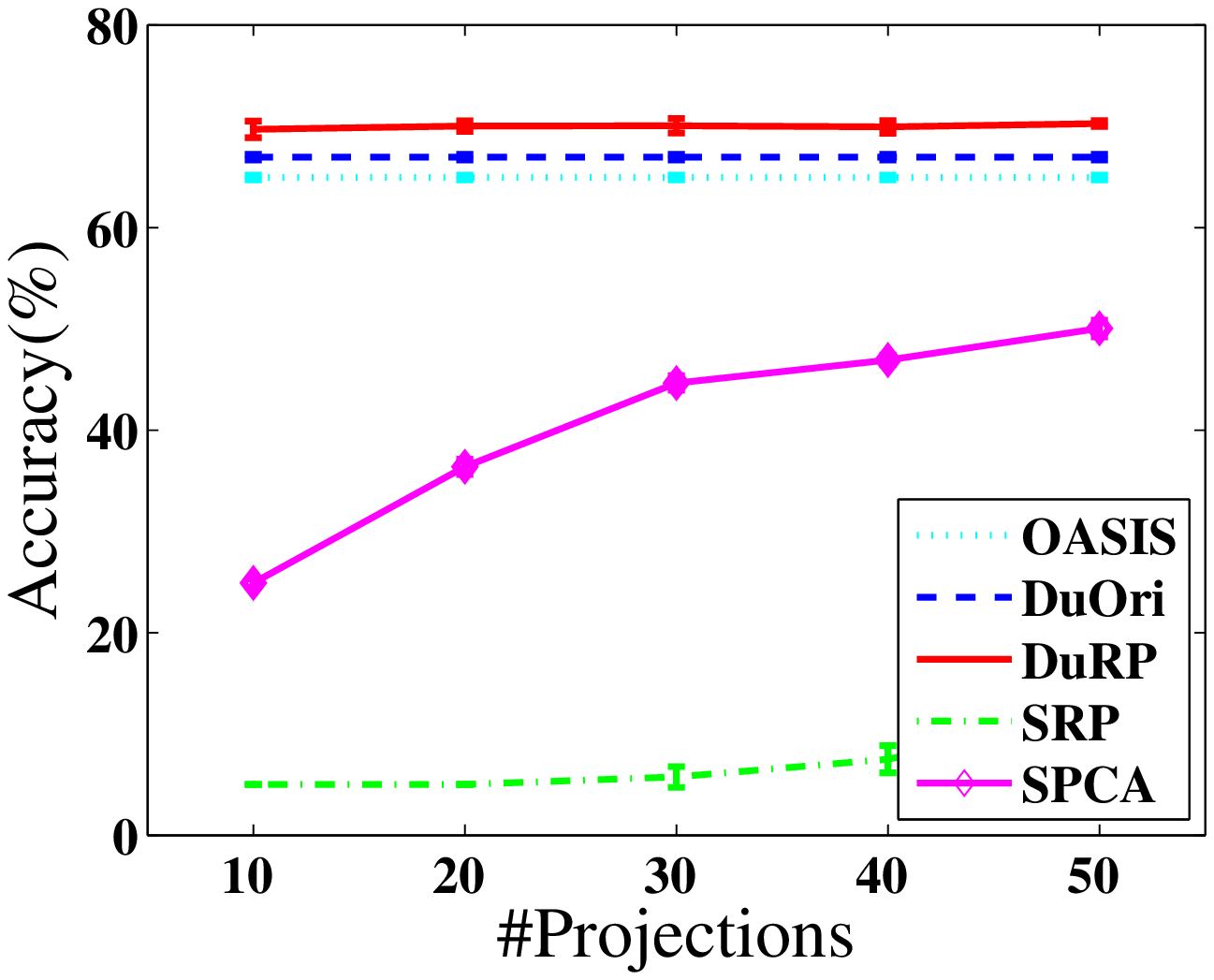}\\
\mbox{\footnotesize (e) {\it 20news}}
\end{minipage}
\begin{minipage}[h]{1.9in}
\centering
\includegraphics[width= 2.0in ]{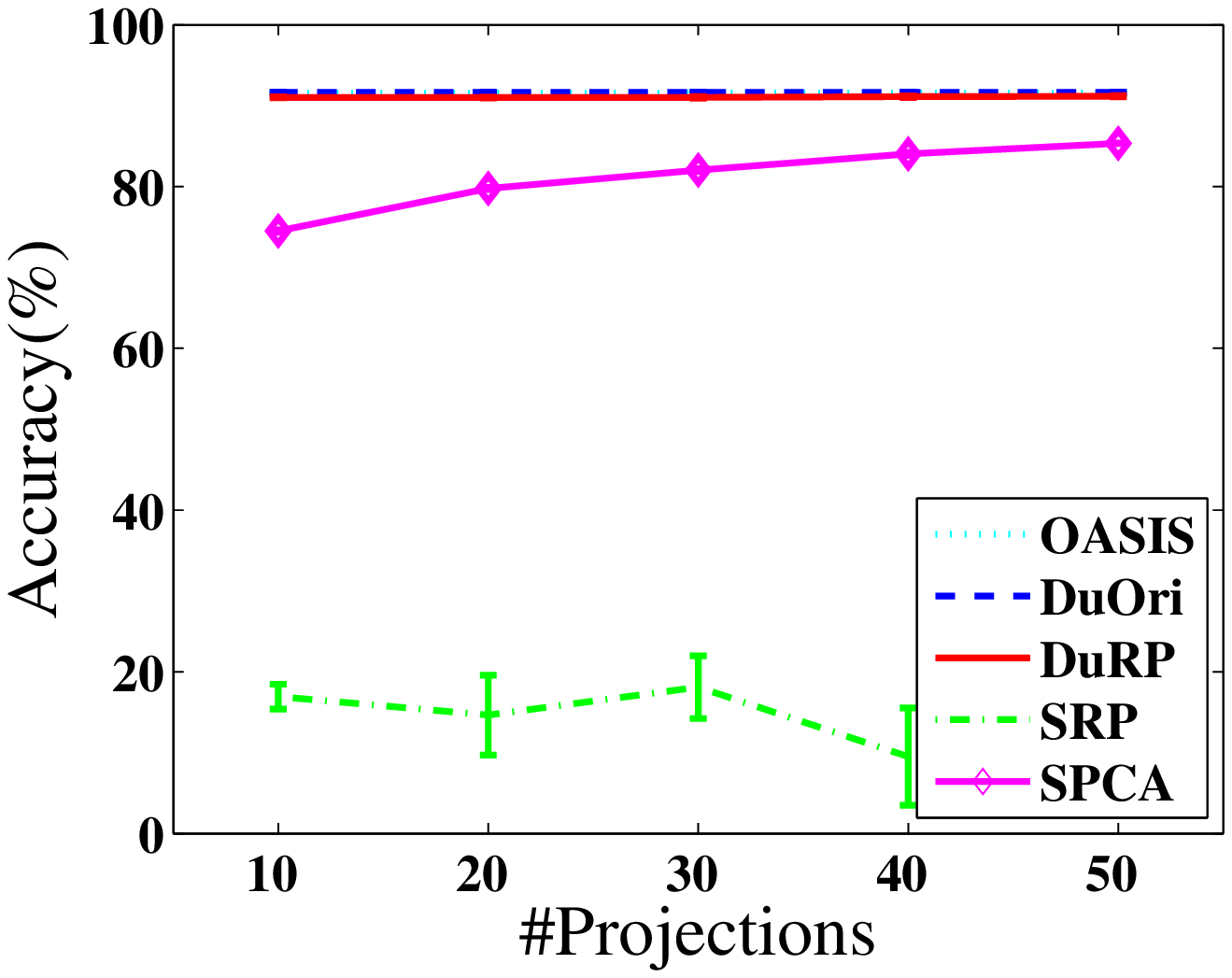}\\
\mbox{\footnotesize (f) {\it rcv30}}
\end{minipage}
\caption{The comparison of different stochastic algorithms for classification}\label{classify}
\end{figure*}

In this experiment, we evaluate the learned metric by its classification accuracy with $k$-NN ($k=5$) classifier. We emphasize that the purpose of this experiment is to evaluate the metrics learned by different DML algorithms, not to demonstrate that the learned metric will result in the state-of-art classification performance\footnote{Many studies (e.g.,~\citep{weinberger2009,Xu2012}) have shown that metric learning do not yield better classification accuracy than the standard classification algorithms (e.g., SVM) given a sufficiently large number of training data.}. Similar to the evaluation by ranking, all experiments are run five times and the results averaged over five trials with standard deviation are reported in Fig.~\ref{classify}. We essentially have the same observation as that for the ranking experiments reported in Section~\ref{sec:exp:rank} except that for most datasets, the three methods DuRP, DuOri, and OASIS yield very similar performance.

Note the main concern of this paper is time efficiency and the size of learned metric is $d\times d$. It is straightforward to store the learned metric efficiently by keeping a low-rank approximation of it.

\section{Conclusion}
\label{sec:conclusion}

In this paper, we propose a dual random projection method to learn the distance metric for large-scale high-dimensional datasets. The main idea is to solve the dual problem in the subspace spanned by random projection, and then recover the distance metric in the original space using the estimated dual variables. We develop the theoretical guarantee that with a high probability, the proposed method can accurately recover the optimal solution with small error when the data matrix is of low rank, and the optimal dual variables even when the data matrix cannot be well approximated by a low rank matrix. Our empirical study confirms both the effectiveness and efficiency of the proposed algorithm for DML by comparing it to the state-of-the-art algorithms for DML. In the future, we plan to  further improve the efficiency of our method by exploiting the scenario when optimal distance metric can be well approximated by a low rank matrix.

\appendix

\section{Proof of Theorem~\ref{thm:low-rank}}

First, we want to prove that $\Gh$ is a good estimation for $G$.
We rewrite $G_{a,b}$ by Kronecker product:
\begin{flalign*}
\lefteqn{G_{a,b} = \langle A_a,A_b\rangle}\\
&\!=\!\langle (\x_i^a\!-\!\x_k^a)(\x_i^a\!-\!\x_k^a)^\top\!-\!(\x_i^a\!-\!\x_j^a)(\x_i^a\!-\!\x_j^a)^\top,(\x_i^b\!-\!\x_k^b)(\x_i^b\!-\!\x_k^b)^\top\!-\!(\x_i^b\!-\!\x_j^b)(\x_i^b\!-\!\x_j^b)^\top\rangle&\\
&\!=\!\langle (\x_i^a\!-\!\x_k^a)\!\otimes\!(\x_i^a\!-\!\x_k^a)\!-\!(\x_i^a\!-\!\x_j^a)\!\otimes\!(\x_i^a\!-\!\x_j^a),(\x_i^b\!-\!\x_k^b)\!\otimes\!(\x_i^b\!-\!\x_k^b)\!-\!(\x_i^b\!-\!\x_j^b)\!\otimes\!(\x_i^b\!-\!\x_j^b)\rangle&\\
&= \langle \z_a, \z_b\rangle&
\end{flalign*}
where $\z_t = (\x_i^t-\x_k^t)\otimes(\x_i^t-\x_k^t)-(\x_i^t-\x_j^t)\otimes(\x_i^t-\x_j^t)$. Define $Z = [\z_1,\cdots,\z_N]$, we have $G = Z^\top Z$.

Under the low rank assumption that all training examples lie in the subspace of $r$-dimension, the dataset $\X = [\x_1,\cdots, \x_n]$ can be decomposed as:
\begin{eqnarray*}
\X = U\Sigma V = \sum_{i=1}^r \lambda_i\u_i\v_i^\top
\end{eqnarray*}
where $\lambda_i$ is the $i$-th singular value of $\X$, and $\u_i$ and $\v_i$ are the corresponding left and right singular vectors of $\X$.
Given the property of Kronecker product that $(A\otimes B)(C\otimes D) = (AC)\otimes(BD)$, we have:
\begin{eqnarray*}
\z_t & = & (\x_i^t - \x_k^t)\otimes(\x_i^t - \x_k^t) - (\x_i^t - \x_j^t)\otimes(\x_i^t - \x_j^t) \\
& = & [U(\xt_i^t - \xt_k^t)]\otimes[U(\xt_i^t - \xt_k^t)] - [U(\xt_i^t - \xt_j^t)]\otimes[U(\xt_i^t - \xt_j^t)] \\
& = & (U\otimes U)\left[(\xt_i^t - \xt_k^t)\otimes(\xt_i^t - \xt_k^t)- (\xt_i^t - \xt_j^t)\otimes(\xt_i^t - \xt_j^t)\right]
\end{eqnarray*}
where $\xt_i^t = U^{\top}\x_i^t$. Define $\Zt = (\zt_1, \ldots, \zt_n)$, where $\zt_t = (\xt_i^t - \xt_j^t)\otimes(\xt_i^t - \xt_j^t) - (\xt_i^t - \xt_k^t)\otimes(\xt_i^t - \xt_k^t)$, we have:
\begin{eqnarray*}
G = \Zt^{\top}(U^{\top}\otimes U^{\top})(U\otimes U)\Zt = \Zt^{\top}(I_r\otimes I_r)\Zt = \Zt^{\top}\Zt
\end{eqnarray*}
where $I_r\otimes I_r$ equals to the identity operator of $r^2\times r^2$.

With the random projection approximation, we have:
\begin{eqnarray*}
\z_t & = & (R^\top(\x_i^t - \x_k^t))\otimes(R^\top(\x_i^t - \x_k^t)) - (R^\top(\x_i^t - \x_j^t))\otimes(R^\top(\x_i^t - \x_j^t)) \\
& = & [R^\top U(\xt_i^t - \xt_k^t)]\otimes[R^\top U(\xt_i^t - \xt_k^t)] - [R^\top U(\xt_i^t - \xt_j^t)]\otimes[R^\top U(\xt_i^t - \xt_j^t)] \\
& = & (R^\top\otimes R^\top)(U\otimes U)\left[(\xt_i^t - \xt_k^t)\otimes(\xt_i^t - \xt_k^t) - (\xt_i^t - \xt_j^t)\otimes(\xt_i^t - \xt_j^t)\right]
\end{eqnarray*}
So,
\begin{eqnarray*}
\Gh& = &\Zt^{\top}(U^{\top}\otimes U^{\top})(RR^\top\otimes RR^\top)(U\otimes U)\Zt \\
&=& \Zt([U^{\top}RR^{\top}U]\otimes[U^{\top}RR^{\top}U])\Zt
\end{eqnarray*}

In order to bound the difference between $G$ and $\Gh$, we need the following corollary:
\begin{cor} \label{cor:gaussian}
~\citep{Zhang13} Let $S \in \R^{r\times m}$ be a standard Gaussian random matrix. Then, for any $0<\varepsilon\leq 1/2$, with a probability $1 - \delta$, we have
\[
    \left\|\frac{1}{m}SS^{\top} - I \right\|_2 \leq \varepsilon
\]
provided
\[
    m \geq \frac{(r+1)\log(2r/\delta)}{c\varepsilon^2}
\]
where constant $c$ is at least $1/4$.
\end{cor}
Define $\Delta = U^{\top}RR^{\top}U - I_r$. Using Corollary.~\ref{cor:gaussian}, with a probability $1 - \delta$, we have $\|\Delta\|_2 \leq \varepsilon$. Using the notation $\Delta$, we have the following expression for $\Gh - G$
\begin{eqnarray*}
\Gh - G & = & \Zt^{\top}\left((I_r + \Delta)\otimes(I_r + \Delta) - I_r\otimes I_r\right)\Zt \\
& = & \Zt^{\top}\left(\Delta\otimes I_r + I_r\otimes \Delta + \Delta \otimes \Delta\right)\Zt = \Zt^{\top}\Gamma\Zt
\end{eqnarray*}
where $\Gamma = \Delta\otimes I_r + I_r\otimes \Delta + \Delta \otimes \Delta$. Using the fact that the eigenvalue values of $A\otimes B$ is given by $\lambda_i(A)\lambda_j(B)$, it is easy to verify that,
\begin{eqnarray*}
\|\Gamma\|_2 \leq \|\Delta\|_2 + \|\Delta\|_2 + \|\Delta\|_2\|\Delta\|_2
\end{eqnarray*}
Using the fact that $\|\Delta\|_2 \leq \varepsilon$ and taking $\e\leq 1/6$ which results in $c\geq 1/3$, with a probability $1 - \delta$, we have
\begin{eqnarray}
\|\Gamma\|_2 \leq 3\varepsilon \label{eqn:bound-1}
\end{eqnarray}

Define $L({\bm \alpha})$ and $\widehat{L}({\bm \alpha})$ as
\begin{eqnarray*}
    &&L({\bm \alpha}) = -\sum_{i=1}^n \ell_*(\alpha_i) - \frac{1}{2\lambda N} {\bm \alpha}^{\top}G{\bm \alpha}, \quad\\
   && \widehat{L}({\bm \alpha}) = -\sum_{i=1}^n \ell_*(\alpha_i) - \frac{1}{2\lambda N} {\bm \alpha}^{\top}\Gh{\bm \alpha}
\end{eqnarray*}

We are now ready to give the proof for Theorem 1. The basic logic is straightforward. Since $\Gh$ is close to $G$, we would expect $\ah_*$, the optimal solution to $\widehat{L}({\bm \alpha})$, to be close to $\as$, the optimal solution to  $L({\bm \alpha})$. Since both $M_*$ and $\M$ are linear in the dual variables $\as$ and $\ah_*$, we would expect $\M$ to be close to $M_*$.

Since $\ah_*$ maximizes $\widehat{L}({\bm \alpha})$ over its domain, which means $(\as-\ah_*)^\top \nabla\widehat{L}(\ah_*)\leq 0$, we have
\begin{eqnarray}
\widehat{L}(\ah_*) \geq \widehat{L}(\as) + \frac{1}{2\lambda N} (\ah_* - \as)^{\top}\Gh(\ah_* - \as) \label{eqn:1}
\end{eqnarray}
Using the concaveness of $\widehat{L}({\bm \alpha})$ and the fact that $\as$ maximizes $L({\bm \alpha})$ over its domain, we have:
\begin{eqnarray}
    &&\widehat{L}(\ah_*)  \leq  \widehat{L}(\as) + (\ah_* - \as)^{\top}\nabla \widehat{L}(\as) - \frac{1}{2\lambda N}(\ah_* - \as)^{\top}\Gh(\ah_* - \as) \nonumber \\
    & = & \widehat{L}(\as) - \frac{1}{2\lambda N}(\ah_* - \as)^{\top}\Gh(\ah_* - \as) + (\ah_* - \as)^{\top}\left(\nabla \widehat{L}(\as) - \nabla L(\as) + \nabla L(\as)\right) \nonumber\\
    & \leq & \widehat{L}(\as) + \frac{1}{\lambda N}(\ah_* - \as)^{\top}(G-\Gh)(\as ) - \frac{1}{2\lambda N}(\ah_* - \as)^{\top}\Gh(\ah_* - \as) \label{eqn:2}
\end{eqnarray}
Combining the inequalities in (\ref{eqn:1}) and (\ref{eqn:2}), we have
\begin{eqnarray*}
\frac{1}{\lambda N}(\ah - \as)^{\top}(G-\Gh )\as \geq \frac{1}{\lambda N}(\ah - \as)^{\top}\Gh(\ah - \as)
\end{eqnarray*}
or
\begin{eqnarray*}
&&( \as- \ah_*)^{\top}(\Gh-G ) \as \geq (\ah_* - \as)^{\top}G(\ah_* - \as)+(\ah_* - \as)^{\top}(\Gh-G)(\ah_* - \as)
\end{eqnarray*}

Define  $\p_* = \Zt \as , \widehat{\p}_* = \Zt \ah_*$, we have:
\begin{eqnarray*}
     (\p_*-\widehat{\p}_*)^\top\Gamma \p_*  \geq \|\widehat{\p}_* - \p_*\|_2^2 + (\widehat{\p}_* - \p_*)^\top\Gamma(\widehat{\p}_* - \p_*)
\end{eqnarray*}
Using the bound given in (\ref{eqn:bound-1}), with a probability $1 - \delta$, we have
\begin{eqnarray*}
\|\widehat{\p}_* - \p_*\|_2 \leq \frac{3\varepsilon}{1 - 3\varepsilon}\|\p_*\|_2
\end{eqnarray*}
We complete the proof by using the fact
\begin{eqnarray*}
\|M_* - \M\|_F = \frac{1}{\lambda N}\|\p_* - \widehat{\p}_*\|_2,\; \|M_*\|_F =\frac{1}{\lambda N} \|\p_*\|_2
\end{eqnarray*}
and~\citep{stark1998vector}
\begin{eqnarray*}
 \|\Pi_{PSD}(M_*)-\Pi_{PSD}(\M)\|_F\leq \|M_* - \M\|_F
\end{eqnarray*}

\section{Proof of Theorem~\ref{thm:dual-recovery}}

Our analysis is based on the following two theorems.
\begin{thm} \label{thm:jl} (Theorem 2~\citep{Blum05}) Let $\x \in \R^d$, and $\xh = R^{\top} \x/\sqrt{m}$, where $R \in \R^{d \times m}$ is a random matrix whose entries are chosen independently from $\N(0, 1)$. Then:
\vspace{-3mm}
\begin{eqnarray*}
    \Pr\left\{(1 - \varepsilon)\|\x\|_2^2 \leq \|\xh\|_2^2 \leq (1 + \varepsilon)\|\x\|_2^2 \right\}\geq 1 - 2\exp\left(-\frac{m}{4}(\varepsilon^2 - \varepsilon^3) \right).
\end{eqnarray*}
\end{thm}

\begin{thm} \label{thm:hardmard} (Lemma B-1~\citep{Karoui10})
Suppose $M$ is a real symmetric matrix with non-negative entries, and $E$ is a real
symmetric matrix such that $\max_{i,j}|E_{i,j}| \leq \xi$. Then, $\|E \circ M \|_2 \leq \xi\|M\|_2$, where $\|\cdot\|_2$ stands for the spectral norm of matrix and $E \circ M$ is the element-wise product between matrices $E$ and $M$.
\end{thm}
Define $L({\bm \alpha})$ and $\widehat{L}({\bm \alpha})$ as
\begin{eqnarray*}
  &&  L({\bm \alpha}) = -\sum_{i=1}^N \ell_*(\alpha_i) - \frac{1}{2\lambda N} {\bm \alpha}^{\top}G{\bm \alpha}, \quad\\
  &&  \widehat{L}({\bm \alpha}) = -\sum_{i=1}^N \ell_*(\alpha_i) - \frac{1}{2\lambda N} {\bm \alpha}^{\top}\Gh{\bm \alpha}
\end{eqnarray*}
Since $\ell(z)$ is $\gamma$-smooth, we have $\ell_*(\alpha)$ be $\gamma^{-1}$-strongly-convex. Using the fact that $\ah_*$ approximately maximizes $\widehat{L}({\bm \alpha})$ with $\eta$-suboptimality and $\ell_*(\cdot)$ is $\gamma^{-1}$-strongly-convex, we have
\begin{eqnarray}
\widehat{L}(\ah_*) \geq \widehat{L}(\as)\! +\! \frac{1}{2\lambda N} (\ah_* - \as)^{\top}(\gamma^{-1} I + \Gh)(\ah_* - \as)\! -\! \eta\label{eqn:1}
\end{eqnarray}
Using the concaveness of $\widehat{L}({\bm \alpha})$ and the fact that $\as$ maximizes $L({\bm \alpha})$ over its domain, we have:
\begin{eqnarray}
    \widehat{L}(\ah_*) \leq \widehat{L}(\as) + \frac{1}{\lambda N}(\ah_* - \as)^{\top}(G-\Gh)\as  - \frac{1}{2\lambda N}(\ah_* - \as)^{\top}(\gamma^{-1} I + \Gh)(\ah_* - \as) \label{eqn:2}
\end{eqnarray}
Combining the inequalities in (\ref{eqn:1}) and (\ref{eqn:2}), we have
\begin{eqnarray*}
\eta + ( \as- \ah_*)^{\top}(\Gh-G ) \as \geq \frac{1}{\gamma}\|\ah_* - \as\|_2^2
\end{eqnarray*}
when we set $\lambda = 1/N$. Using the fact $(\as - \ah_*)(\Gh - G) \as \leq \|\as - \ah_*\|_2\|\as\|_2\|\Gh - G\|_2$, we have
\begin{eqnarray}
\|\as - \ah_*\|^2_2 \leq \gamma \|\as - \ah_*\|_2\|\as\|_2\|\Gh - G\|_2 + \gamma\eta , \nonumber
\end{eqnarray}
implying that
\begin{eqnarray}
\|\as - \ah_*\|_2 \leq \max\left(2\gamma \|\Gh - G\|_2 \|\as\|_2, \sqrt{2\gamma\eta}\right). \label{eqn:bound-temp}
\end{eqnarray}

To bound $\|\as - \ah_*\|_2$, we need to bound $\|\Gh - G\|_2$. To this end, we write the $G_{a, b}$ as
\begin{eqnarray*}
&&G_{a,b} = \langle A_a,A_b\rangle\\
&=&\left[(\x_i^a \!-\! \x_k^a)^{\top}(\x_i^b \!-\! \x_k^b)\right]^2 \!+\! \left[(\x_i^a \!-\! \x_j^a)^{\top}(\x_i^b \!-\! \x_j^b)\right]^2 \\
&-& \left[(\x_i^a \!-\! \x_k^a)^{\top}(\x_i^b \!-\! \x_j^b)\right]^2 \!-\! \left[(\x_i^a \!-\! \x_j^a)^{\top}(\x_i^b \!-\! \x_k^b)\right]^2
\end{eqnarray*}
Similarly, we write $\Gh_{a,b}$ as
{\small
\begin{eqnarray*}
&&\Gh_{a,b} = \langle R^{\top}A_aR,R^{\top}A_bR\rangle\\
\!\!\!\!&\!\!\!=\!\!\!&\!\!\!\left[(\x_i^a\! -\! \x_k^a)^{\top}RR^{\top}(\x_i^b \!-\! \x_k^b)\right]^2 \!\!+\!\! \left[(\x_i^a \!-\! \x_j^a)^{\top}RR^{\top}(\x_i^b \!-\! \x_j^b)\right]^2 \\
\!\!\!\!&\!\!\!-\!\!\!&\!\!\! \left[(\x_i^a \!-\! \x_k^a)^{\top}RR^{\top}(\x_i^b \!-\! \x_j^b)\right]^2\!\!-\!\! \left[(\x_i^a \!-\! \x_j^a)^{\top}RR^{\top}(\x_i^b \!-\! \x_k^b)\right]^2
\end{eqnarray*}}
Hence, we can write $\Gh - G = B^1 + B^2 + B^3 + B^4$, where $B^1$, $B^2$, $B^3$, and $B^4$ are defined as
\begin{eqnarray*}
&&B^1_{a,b} = \!\! \left[(\x_i^a \!-\! \x_k^a)^{\top}RR^{\top}(\x_i^b \!-\! \x_k^b)\right]^2 \!-\! \left[(\x_i^a \!-\! \x_k^a)^{\top}(\x_i^b \!-\! \x_k^b)\right]^2 \\
&&B^2_{a,b} = \!\! \left[(\x_i^a \!-\! \x_j^a)^{\top}RR^{\top}(\x_i^b \!-\! \x_j^b)\right]^2 \!-\! \left[(\x_i^a \!-\! \x_j^a)^{\top}(\x_i^b \!-\! \x_j^b)\right]^2 \\
&&B^3_{a,b} = \!\! \left[(\x_i^a \!-\! \x_k^a)^{\top}(\x_i^b \!-\! \x_j^b)\right]^2 \!-\! \left[(\x_i^a \!-\! \x_k^a)^{\top}RR^{\top}(\x_i^b \!-\! \x_j^b)\right]^2 \\
&&B^4_{a,b} = \!\! \left[(\x_i^a \!-\! \x_j^a)^{\top}(\x_i^b \!-\! \x_k^b)\right]^2 \!-\! \left[(\x_i^a \!-\! \x_j^a)^{\top}RR^{\top}(\x_i^b \!-\! \x_k^b)\right]^2
\end{eqnarray*}
Using the result from Theorem~\ref{thm:jl} and the definition of matrices $M^1$, $M^2$, $M^3$, $M^4$, we have, with a probability $1 - \delta$, for any $a, b$,
\[
|B^i_{a,b}| \leq \epsilon M^i_{a,b},\ i = \{1,2,3,4\}
\]
provided that $\epsilon \leq 1/2$ and
\begin{eqnarray}
m \geq \frac{8}{\epsilon^2}\ln\frac{8N}{\delta} \label{eqn:cond-1}
\end{eqnarray}
Using Theorem~\ref{thm:hardmard}, under the condition in (\ref{eqn:cond-1}), we have, with a probability $1 - \delta$,
\begin{eqnarray*}
\|\Gh \!-\! G\|_2 \!\leq\! \epsilon\left(\|M^1\| \!+\! \|M^2\| \!+\! \|M^3\| \!+\! \|M^4\|\right) \!\leq\! 4\kappa\epsilon
\end{eqnarray*}
where the last step uses the definition of $\kappa$. We complete the proof by plugging the bound for $\|\Gh - G\|_2$ into (\ref{eqn:bound-temp}).

\vskip 0.2in

\bibliographystyle{natbib}
\bibliography{dualrp}

\end{document}